%% file: main.tex
\documentclass[3p]{elsarticle}
\journal{Journal of Neural Networks}

\makeatletter
\def\ps@pprintTitle{%
 \let\@oddhead\@empty
 \let\@evenhead\@empty
 \def\@oddfoot{\footnotesize\itshape
       Published in the \ifx\@journal\@empty Elsevier
       \else\@journal\fi\hfill} 
 \let\@evenfoot\@oddfoot}
\makeatother

\usepackage{amsmath}
\usepackage{lineno,hyperref}
\modulolinenumbers[1]
\usepackage{lipsum}
\usepackage{amsfonts}
\usepackage{graphicx}
\usepackage{subcaption}
\usepackage{epstopdf}
\usepackage{algorithmic}
\usepackage{soul}
\usepackage{float}
\usepackage{multirow}
\usepackage{booktabs}
\usepackage{amsbsy}
\usepackage{amssymb}
\usepackage{amsmath} 
\usepackage{amsthm}
\usepackage{algorithmic}
\usepackage{algorithm}
\usepackage{mathtools}
\usepackage{subcaption}
\usepackage{soul}
\usepackage{color}
\usepackage{xcolor}
\usepackage{lineno}
\usepackage{ulem}
\graphicspath{{figures}/}
\usepackage{booktabs}
\usepackage{bm}
\newtheorem{rmk}{Remark}
\DeclareMathOperator*{\argmin}{arg\,min}

\theoremstyle{plain}
\newtheorem{theorem}{Theorem}
\newtheorem{definition}[theorem]{Definition}

\begin{document}

\begin{frontmatter}

\title{Kolmogorov n-Widths for Multitask Physics-Informed Machine Learning (PIML) Methods: Towards Robust Metrics}
\author{Michael Penwarden$^{1,2}$, Houman Owhadi$^{3}$, Robert M. Kirby$^{1,2}$}
\cortext[mycorrespondingauthor]{Email addresses: 
 mpenwarden@sci.utah.edu (Michael Penwarden), owhadi@caltech.edu (Houman Owhadi), kirby@cs.utah.edu (Robert M. Kirby)}

\address{$^1$~ Scientific Computing and Imaging Institute, University of Utah, Salt Lake City, UT 84112, USA.}
\address{$^2$~ Kahlert School of Computing, University of Utah, Salt Lake City, UT 84112, USA.}
\address{$^3$~ Department of Computing and Mathematical Sciences, Caltech, Pasadena, CA 91125, USA}

\begin{abstract}
\input{abstract}
\end{abstract}


\end{frontmatter}


\input{introduction}

\input{background}

\input{methods}

\input{results}

\input{summary}

\vspace{0.2in}
\noindent {\bf Acknowledgements:}
This work was funded under AFOSR MURI FA9550-20-1-0358. We would also like to thank Professors Sarang Joshi (University of Utah) and Jerome Darbon (Brown University) for their insightful comments.

\appendix 
\input{appendix}

\newpage
\bibliography{references}

\end{document}

%% file: abstract.tex
Physics-informed machine learning (PIML) as a means of solving partial differential equations (PDEs) has garnered much attention in the Computational Science and Engineering (CS\&E) world. This topic encompasses a broad array of methods and models aimed at solving a single or a collection of PDE problems, called multitask learning. PIML is characterized by the incorporation of physical laws into the training process of machine learning models in lieu of large data when solving PDE problems. Despite the overall success of this collection of methods, it remains incredibly difficult to analyze, benchmark, and generally compare one approach to another. Using Kolmogorov n-widths as a measure of effectiveness of approximating functions, we judiciously apply this metric in the comparison of various multitask PIML architectures. We compute lower accuracy bounds and analyze the model's learned basis functions on various PDE problems. This is the first objective metric for comparing multitask PIML architectures and helps remove uncertainty in model validation from selective sampling and overfitting. We also identify avenues of improvement for model architectures, such as the choice of activation function, which can drastically affect model generalization to ``worst-case'' scenarios, which is not observed when reporting task-specific errors. We also incorporate this metric into the optimization process through regularization, which improves the models' generalizability over the multitask PDE problem.

%% file: introduction.tex
\section{Introduction}
\label{sec:introduction}
Physics-informed machine learning (PIML) has emerged as a popular framework for solving partial differential equations (PDEs). PIML has been particularly effective and efficient in solving ill-posed and inverse problems over traditional methods \cite{Karniadakis2021}. The most ubiquitously used PIML implementation at present is the physics-informed neural network (PINN)\cite{raissi2019physics}. This approach is often selected due to its flexibility in discretization and has been shown to be successful across a wide class of application domains \cite{mathews2021uncovering,kissas2020machine,10.1371/journal.pcbi.1007575,WANG2021109914,shukla2020physics,Chen:20,10.3389/fphy.2020.00042}. Innumerable strategies to enhance PINN training have been proposed such as adaptive sampling \cite{lu2021deepxde,daw2022rethinking,subramanian2022adaptive}, adaptive weighting \cite{wang2022respecting,mcclenny2020self}, adaptive activation functions \cite{jagtap2020adaptive,jagtap2022important}, additional loss terms \cite{yu2022gradient}, domain decomposition \cite{JagtapK,JAGTAP2020113028,meng2020ppinn,hu2022augmented}, metalearning \cite{PENWARDEN2023111912,li2022meta,psaros2022meta}, and network architecture modification to obey characteristics \cite{mojgani2022lagrangian, braga2022characteristics}. A thorough summary of PINN training challenges and their proposed solutions is provided in \cite{mojgani2022lagrangian}. The physics-informed concept has also been incorporated into neural operator learning in works such as physics-informed DeepONets \cite{wang2021learning,goswami2022physics,goswami2022physics2} and physics-informed neural operators (PINO) \cite{li2021physics,Konuk2021PhysicsguidedDL}. Comprehensive surveys of the state of PIML and future directions can be found in \cite{cuomo2022scientific,hao2022physics}.

The optimization process of PINNs and physics-informed methods, in general, limits the upper accuracy bound and can cause training challenges due to the interplay between minimizing the PDE residuals and the non-convex optimization of deep neural networks, which induces non-unique solutions \cite{PENWARDEN2023112464}. Training challenges in PINNs can happen for various reasons, the most common being poor sampling, unbalanced loss term weights, or poor optimization schemes. Even with a well-tuned PINN, ``stiff" PDEs with sharp transitions \cite{wang2021understanding}, multiscale problems \cite{wang2021eigenvector}, or highly nonlinear time-varying PDEs \cite{MATTEY2022114474} can still pose problems for the standard PINN. Despite the overall success of these methods, many practical issues and avenues of improvement still exist, particularly in the case of forward problems \cite{WANG2022110768,wang2022respecting}. Many works still handpick problems or tasks in a family of problems that show the most promising results and do not provide a lower accuracy bound or performance over the continuous task family. 
Therefore, we propose using Kolmogorov n-widths applied to physics-informed methods as a measure with which we can analyze, benchmark, and improve models for multitask problems. Other connections between Kolmogorov n-widths and PINNs have been proposed as a measure of the rate of convergence \cite{mojgani2022lagrangian}; however, only approximations using the rate of decay of temporal snapshot singular values are made as opposed to the true \textit{inf sup inf} evaluation performed here. 

Our first major contribution is a proposed methodology for computing Kolmogorov n-widths as a metric and regularizer for PIML models, which requires tri-optimization. Our second contribution is an experimental study of different multitask PIML architectures for which we can compute benchmarks to compare against. We also provide insights into the ability of different architectures to learn generalizable basis functions through choice in various parameters, such as the activation function. The paper is organized as follows: in Section \ref{sec:background}, we first summarize PIML and existing architectures that target multitask PDE problems such as physics-informed DeepONets and multihead PINNs. In Section \ref{sec:methods}, we introduce Kolmogorov n-widths and how they can be numerically approximated for PIML models using a novel optimization procedure. Two applications of this metric are proposed: the first as an analysis tool with which we can benchmark different architectures and compute lower accuracy bounds, and the second is a way to add regularization into training to improve the lower accuracy bound and the generalization of the learnable neural network basis functions. In Section \ref{sec:numerical_results}, we provide numerical results on various multitask problems, study the effect of different network sizes and training iterations, and study the effect of regularization on the learned basis functions. We summarize and conclude our results in Section \ref{sec:conclusions}.

%% file: background.tex
\section{Background}
\label{sec:background}
\begin{figure}
\centering
\includegraphics[width=0.9\textwidth]{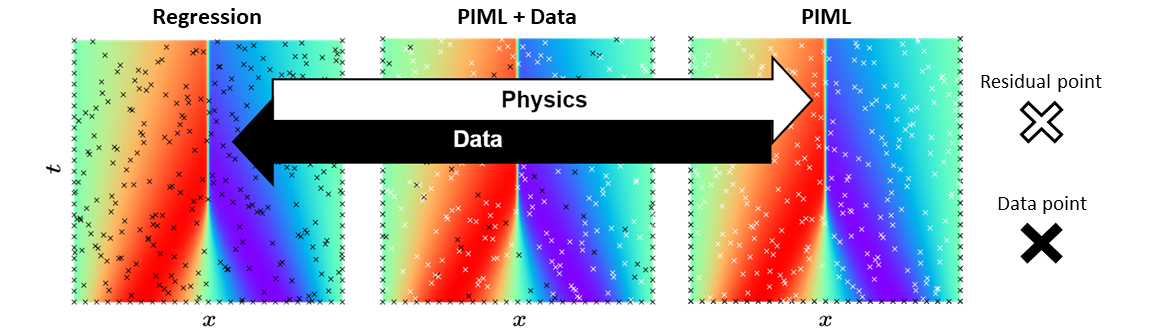}
  \caption{Physics-Informed Machine Learning as a spectrum of data and physics. Residual points are where the PDE residual is evaluated and minimized during optimization.}
  \label{fig:PIML}
\end{figure}

Physics-informed machine learning (PIML) \cite{Karniadakis2021,cuomo2022scientific,hao2022physics} was introduced as an alternative to traditional PDE solvers such as finite differences, finite elements, finite volumes, spectral methods, etc. Despite significant efforts with traditional approaches, it is still difficult to incorporate noisy data, solve high-dimensional and inverse problems, or overcome great costs attributed to complex mesh generation, all requiring elaborate code to function. Alternatively, PIML represents the modern confluence of two powerful computational modeling paradigms: data-intensive machine learning and model-informed simulation science. Using machine learning as the foundation, this approach benefits from the rich expressivity of deep neural networks (DNNs) as a surrogate for PDE solutions. They also allow for “meshfree” discretization and capitalize on machine learning technologies such as automatic differentiation and stochastic optimization to easily incorporate physical laws with data \cite{DeepLearning}. Figure \ref{fig:PIML} illustrates the spectrum of situations PIML can be placed in, ranging from pure physics-informed learning to pure regression. This work focuses solely on the physics-informed paradigm for multitask forward problems since they result in challenging training cases and compete with dataless traditional PDE solvers \cite{krishnapriyan2021characterizing,PENWARDEN2023112464}.

\subsection{Physics-Informed Neural Networks}
\begin{figure}
\centering
\includegraphics[width=0.9\textwidth]{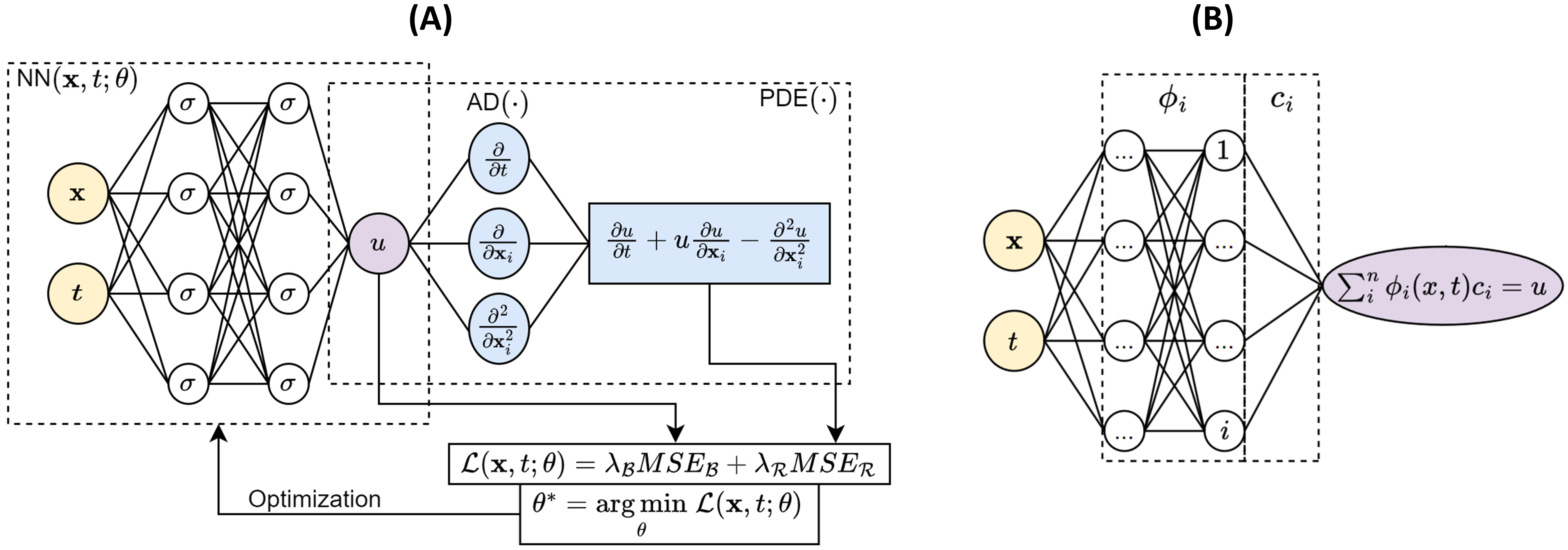}
  \caption{(A) Diagram of full Physics-Informed Neural Network including the PDE residual formulation using automatic differentiation and optimization process. (B) PINN solution as a sum of basis functions ($\phi_i$) and coefficients ($c_i$).}
  \label{fig:PINN_diagram}
\end{figure}

Physics-informed neural networks (PINNs), proposed in \cite{raissi2019physics,raissi2017physicsI,raissi2017physicsII}, are the most common paradigm of PIML. The NN component of PINN is generally taken as a feed-forward NN (FF-NN), which is the case in this work, but includes and is not limited to convolutional NN (CNN) \cite{GAO2021110079,9403414,ZHU201956,GENEVA2020109056,WANG2021114037}, recurrent NN (RNN) \cite{VIANA2021106458,ZHANG2020113226,YUCESAN2021103386}, among others. In the original PINNs work, when presented with a PDE specified over a spatiotemporal domain $\Omega \times T$, the solution is computed (i.e., the differential operator is satisfied) as in other mesh-free methods like RBF-FD \cite{ShankarWFK1,ShankarWFK2} at a collection of collocation points. To perform this optimization, we re-write a generalized PDE system in residual form as $\mathcal{R}(u) = \mathcal{S} - \frac{\partial}{\partial t} u - \mathcal{F}(u)$ where $\mathcal{S}$ and $\mathcal{F}$ are source functions and nonlinear operators respectively and compute the derivatives with automatic-differentiation (AD). PINN optimization is expressed as follows: given a neural network $u_{\bm{\theta}}(\mathbf{x},t)$ with specified activation functions and learnable parameters $\bm{\theta}$, find $\bm{\theta}^*$ that minimizes the summed error of the boundary/initial condition and PDE residual term. The PINN optimization problem is explicitly defined as follows:
\begin{align}
\label{eq:PINN}
\bm{\theta}^* \gets \argmin_{\bm{\theta}} \frac{\lambda_{\mathcal{B}}}{N} \sum^{N}_{j=1} \left| u(\mathbf{x}_{\mathcal{B}_j},t_{\mathcal{B}_j}) - u_{\bm{\theta}}(\mathbf{x}_{\mathcal{B}_j},t_{\mathcal{B}_j}) \right|^{2} + \frac{\lambda_{\mathcal{R}}}{M} \sum^{M}_{k=1} \left| \mathcal{R} \left( u_{\bm{\theta}}(\mathbf{x}_{\mathcal{R}_k},t_{\mathcal{R}_k}) \right) \right|^{2},
\end{align}
where $\{ \mathbf{x}_{\mathcal{B}},t_{\mathcal{B}} \}$ denote the initial and boundary condition point set and  $\{ \mathbf{x}_{\mathcal{R}},t_{\mathcal{R}} \}$ specify the collocation points for evaluation of the collocating residual term $\mathcal{R}(u_{\bm{\theta}})$. Given the optimization scheme in Equation \ref{eq:PINN}, we now draw the parallel of neural network architecture to the linear combination of global basis functions. In Figure \ref{fig:PINN_diagram}, diagram (A) outlines the framework for a standard PINN with physics-informed training as described in Equation \ref{eq:PINN}. In diagram (B) and future diagrams, physics-informed training is implied but not explicitly shown. In (B), we see that the ``body'' of the neural network acts as a surrogate for learning global basis functions, where each neuron in the final hidden layer is one basis function, similar to a set of global basis functions in spectral methods. This is a unique and overlooked aspect of PINNs, which are commonly compared to traditional methods that use basis functions with compact support instead of global functions. The final layer of the network can be viewed as the basis coefficients, which are solution-specific.

\subsection{Multitask PDE Problems}
Multitask learning \cite{caruana1997multitask} is when multiple tasks are solved simultaneously while exploiting commonalities across the tasks. This learning paradigm improves generalization by incorporating domain information contained in \textit{related} tasks as an inductive bias. In the context of PDEs, this can arise in many forms, such as solving a task family in which the initial condition, boundary condition, forcing or source functions, PDE parameters, etc., change. Examples of use cases for multitask PDE problems for PINNs can be found in \cite{PENWARDEN2023111912,penwarden2022multifidelity,bahmani2021training,thanasutives2021adversarial,wang2023training}. PINNs themselves are not a multitask model without modification since they each solve one task individually. In this work, we are primarily interested in the representation of learned global basis functions in multitask PIML models arising from solving various multitask PDE problems and how well they generalize to difficult unseen tasks.

\subsection{Multihead PINNs}
\begin{figure}
\centering
\includegraphics[width=0.8\textwidth]{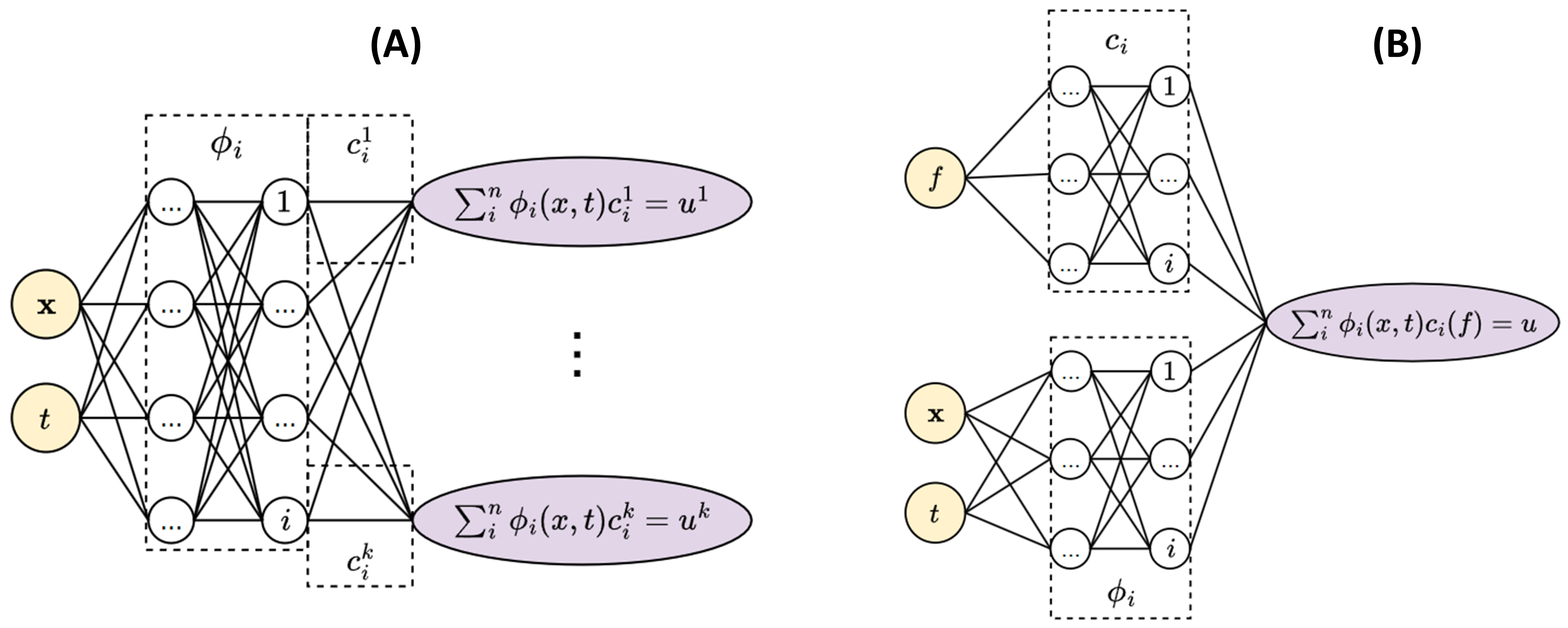}
  \caption{(A) Multihead PINN where each ``head'' is a different linear combination ($c_i$) of the body network basis functions ($\phi_i$). (B) Physics-Informed DeepONet architecture where the ``branch'' network represents the coefficients ($c_i$) and the ``trunk'' network represents the basis functions ($\phi_i$).}
  \label{fig:MHPINN+PIDON_diagram}
\end{figure}

Multihead PINNs (MH-PINNs) were introduced in \cite{zou2023hydra} as a framework for solving multiple PDE problems with PINNs without having to train $K$ number of independent networks. This modification makes MH-PINN a potent tool for multitask learning, generative modeling, and few-shot learning. Given this modification, we can modify the PINN optimization problem in Equation \ref{eq:PINN} as the new problem in Equation \ref{eq:MHPINN}. The primary difference is that MH-PINNs have an outer sum over all tasks $K$ and that each head creates a unique network prediction indexed as $u_{\bm{\theta}}^{(i)}$ for each task. 
\begin{equation}
\begin{aligned}
\label{eq:MHPINN}
\bm{\theta}^* \gets \argmin_{\bm{\theta}} \sum^{K}_{i=1} \Biggl( \frac{\lambda_{\mathcal{B}}}{N} \sum^{N}_{j=1} \left| u^{(i)}(\mathbf{x}_{\mathcal{B}_j},t_{\mathcal{B}_j}) - u^{(i)}_{\bm{\theta}}(\mathbf{x}_{\mathcal{B}_j},t_{\mathcal{B}_j}) \right|^{2} + 
\frac{\lambda_{\mathcal{R}}}{M} \sum^{M}_{k=1} \left| \mathcal{R} \left( u^{(i)}_{\bm{\theta}}(\mathbf{x}_{\mathcal{R}_k},t_{\mathcal{R}_k}) \right) \right|^{2} \Biggr)
\end{aligned}
\end{equation}
In Figure \ref{fig:MHPINN+PIDON_diagram}, we denote two types of multitask PIML models. The first model (A) is one in which the ``body'' represents the global basis function, and the head represents the task-specific operations. Given this framework, a shared basis for the multitask problem is learned, and optimal coefficients are found to solve different problems, similar to spectral methods. Therefore, the problem is: how can we evaluate if our model has learned an optimal basis over the entire family of multitask problems, given it trains on a discrete set? Unlike in spectral methods, we perform complex optimization on a highly expressive neural network to form the set of basis functions instead of using predefined orthogonal ones such as Fourier series or Chebyshev polynomials, making this evaluation difficult. In our proposed methodology, we will return to this concept with Kolmogorov n-widths in Section \ref{sec:methods}.

\subsection{PI-DeepONets}
Deep Operator Networks (DeepONets) are a novel paradigm of scientific machine learning (SciML) in which a neural network is used as a surrogate model for function-to-function operators \cite{lu2021learning}. In the original formulation, the DeepONet performs nonlinear regression between two functions using experimental data. In PDE terms, an example would be learning a mapping between a forcing function and the resulting solution, $\mathcal{G}:f \rightarrow u$, for a fixed PDE and domain. A standard DeepONet consists of two networks: the branch, which represents basis coefficients conditioned on the input function, and the trunk, which represents the basis functions conditioned upon the output location in space and time. This framework, shown in Figure \ref{fig:MHPINN+PIDON_diagram} (B), is based on the universal approximation theorem for operators \cite{392253}.

Operator learning can be beneficial compared to models such as PINNs, which require a new network instance to be optimized for each problem. Neural operators predict new solutions at test time without the need for additional training. More recently, physics-informed DeepONets (PI-DONs) were introduced in \cite{wang2021learning,goswami2022physics,goswami2022physics2} as a way to offset the large data requirement of this type of model. This optimization problem, stated in Equation \ref{eq:PIDON}, is similar to MH-PINNs with the distinction that our output is also conditioned upon the input function $f(\cdot)$ as well as the spatiotemporal coordinates ($\mathbf{x},t$). Both models seek to learn an optimal set of global basis functions for the multitask PDE problem. We, therefore, propose Kolmogorov n-widths as a concrete metric of comparison for models of this physics-informed multitask paradigm. 
\begin{equation}
\begin{aligned}
\label{eq:PIDON}
\bm{\theta}^* \gets \argmin_{\bm{\theta}} \sum^{K}_{i=1} \Biggl( \frac{\lambda_{\mathcal{B}}}{N} \sum^{N}_{j=1} \left| u^{(i)}(\mathbf{x}_{\mathcal{B}_j},t_{\mathcal{B}_j}) - u^{(i)}_{\bm{\theta}}(\mathbf{x}_{\mathcal{B}_j},t_{\mathcal{B}_j}, f(\cdot)) \right|^{2} + 
\frac{\lambda_{\mathcal{R}}}{M} \sum^{M}_{k=1} \left| \mathcal{R} \left( u^{(i)}_{\bm{\theta}}(\mathbf{x}_{\mathcal{R}_k},t_{\mathcal{R}_k}, f(\cdot)) \right) \right|^{2} \Biggr)
\end{aligned}
\end{equation}

%% file: methods.tex
\section{Methods}
\label{sec:methods}

\subsection{Kolmogorov n-widths}
\label{ssec:Knw}
In approximation theory, Kolmogorov n-widths are a measure of effectiveness of approximating functions. In other words, it is a measure of how well any n-dimensional subspace can approximate the solution manifold, defined in \cite{Kolmogoroff1936UberDB, Pinkus1985nWidthsIA, EVANS20091726}. 
\begin{definition}
Let $\mathcal{M}_n$ be any n-dimensional subspace of the normed linear space $\mathcal{M}$ with norm $||\cdot||_{\mathcal{M}}$. For each $x \in \mathcal{M}$,
\begin{align}
\label{eq:distance}
    \delta(\mathcal{M}_n,x) = \inf_{y_n \in \mathcal{M}_n}||x-y_n||_\mathcal{M}
\end{align}
is the distance of $\mathcal{M}_n$ to x, shown in Figure \ref{fig:n-width} (A). If $\delta(\mathcal{M}_n,x) = ||x-y_n^*||_\mathcal{M}$, then $y_n^*$ is the ``best'' approximation from $\mathcal{M}_n$ to $x$. However, we are more often interested in how well an n-dimensional subspace $\mathcal{M}_n$ approximates another subset $\mathcal{A} \subset \mathcal{M}$, such as a solution manifold to a multitask PDE problem. In this case, we define
\begin{align*}
\delta(\mathcal{M}_n,\mathcal{A}) = \sup_{x \in \mathcal{A}} \inf_{y_n \in \mathcal{M}_n} ||x-y_n||_{\mathcal{M}}
\end{align*}
as the deviation from $\mathcal{M}_n$ to $\mathcal{A}$, shown in Figure \ref{fig:n-width} (B). 

\begin{figure}[H]
\centering
\includegraphics[width=0.9\textwidth]{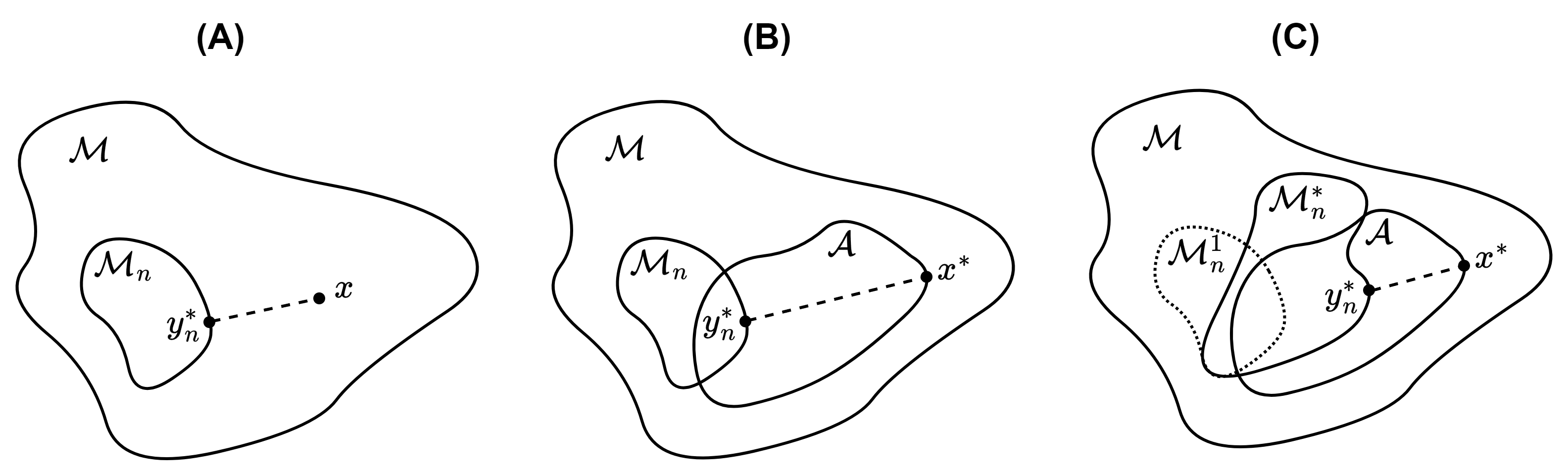}
  \caption{(A) Distance of $\mathcal{M}_n$ to point $x \in \mathcal{M}$ where $y_n^*$ is the best approximation, i.e., satisfies Equation \ref{eq:distance}. (B) Distance of $\mathcal{M}_n$ to $\mathcal{A}$ where $y_n^*$ and $x^*$ achieve the \textit{sup inf} of $\mathcal{M}_n$ to $\mathcal{A}$, i.e., satisfy $\sup_{x \in \mathcal{A}} \inf_{y_n \in \mathcal{M}_n} ||x-y_n||_{\mathcal{M}}$. (C) Distance of $\mathcal{M}^*_n$ to $\mathcal{A}$ where $y_n^*$ and $x^*$ achieve the \textit{inf sup inf} of $\mathcal{M}^*_n$ to $\mathcal{A}$, i.e., satisfy $\inf_{\mathcal{M}_n \in \mathcal{M}} \sup_{x \in \mathcal{A}} \inf_{y_n \in \mathcal{M}_n} ||x-y_n||_{\mathcal{M}}$.}
  \label{fig:n-width}
\end{figure}

Similarly, we can say that if $\delta(\mathcal{M}_n,\mathcal{A}) = ||x^* - y_n^*||_{\mathcal{M}}=\delta(\mathcal{M}_n,x^*)$, then $y_n^*$ is the ``best'' approximation given by $\mathcal{M}_n$ to the ``worst'' or most challenging instance of $x \in \mathcal{A}$ for $\mathcal{M}_n$ to approximate, which is $x^*$. Given two n-dimensional subspaces $\mathcal{M}_n$ and $\mathcal{N}_n$, we can consider the ratio 
\begin{align*}
\xi(\mathcal{A},\mathcal{M}_n,\mathcal{N}_n) = \ \frac{\delta(\mathcal{M}_n,\mathcal{A})}{\delta(\mathcal{N}_n,\mathcal{A})}
\end{align*}
as a metric for comparison when considering which approximating space we prefer. However, again, we are often more interested in how well our approximation $\mathcal{M}_n$ compares to other viable n-dimensional subspaces in $\mathcal{M}$, not just one. Therefore, let us define the Kolmogorov n-width as 
\begin{align}
\label{eq:KnW_original}
\mathcal{K}(\mathcal{M},\mathcal{A}) = \inf_{\mathcal{M}_n \in \mathcal{M}} \sup_{x \in \mathcal{A}} \inf_{y_n \in \mathcal{M}_n} ||x-y_n||_{\mathcal{M}}
\end{align}
which is the deviation from $\mathcal{M}^*_n$ to $\mathcal{A}$, shown in Figure \ref{fig:n-width} (C).
\end{definition}

An n-dimensional subspace $\mathcal{M}_n^*$ is ``optimal'' if $\delta(\mathcal{M}_n^*,\mathcal{A}) = \mathcal{K}(\mathcal{M},\mathcal{A})$. However, it is difficult to obtain $\mathcal{K}(\mathcal{M},\mathcal{A})$ and determine the existence of optional approximating functions.

In terms of PIML models, let us separate our learnable model parameters ($\boldsymbol{\theta}$) into two groups: the parameterization of the basis $\phi(W^1)$ and the parameterization of the coefficients $c(W^2)$, shown in Figures \ref{fig:PINN_diagram} \& \ref{fig:MHPINN+PIDON_diagram} for different architectures. Next, consider the generalized PDE problem
\begin{equation}
\begin{aligned}
\label{eq:PDE_problem}
\mathcal{F}u(x) &= f(x) &&\\
\text{where } u(x; c) &= \sum^{N}_{i=1} c_i \phi_i (x). &&
\end{aligned}
\end{equation}
Note that here $\phi_i(x)$ is known whereas $\phi_i(x, W^1)$ is to be learned by the network. For this prototypical PDE problem where the boundary conditions are enforced by the basis functions $\phi_i$, we can define the PIML Kolmogorov n-width as
\begin{align}
\label{eq:Knw}
\tilde{\mathcal{K}}(\mathcal{M}, \mathcal{A}) =\inf_{W^1} \sup_{c} \inf_{W^2} \left| \left|u(x; \; c) - \tilde{u}(x; \; W^1, W^2)\right| \right|_\mathcal{M}, \; c_i \in [a,b],
\end{align}

\noindent where $\mathcal{A}$ represents the solution manifold to the multitask PDE problem in Equation \ref{eq:PDE_problem} and the \textit{inf} \textit{sup} \textit{inf} is the same form as Equation \ref{eq:KnW_original} but with respect to neural network parameterization. Colloquially, this metric for PIML multitask problems can be described as an error estimate for the best learnable model basis $(\inf_{W^1})$ on the most challenging problem in the bounded multitask solution space $(\sup_{c})$ with the best basis coefficients $(\inf_{W^2})$. This is also the lower accuracy bound for a given architecture. Note that this will be an approximation, as the Kolmogorov n-width is difficult to analytically compute, even in the non-neural network case \cite{EVANS20091726}. Furthermore, Equation \ref{eq:Knw} is linear in $W^1$ and nonlinear in $W^2$ for non-single-layer networks. The nonlinear aspect of this optimization will be addressed in the subsequent section.

\subsection{Multitask PIML Metric}
To compute the Kolmogorov n-width metric, we perform a two-step optimization process. First, we perform standard model optimization for the given multitask PIML model, assuming we provide a sufficient sampling of problems in the multitask PDE space over which to train. This optimization step is explicitly defined in Equation \ref{eq:MHPINN} for multihead PINNs and in Equation \ref{eq:PIDON} for physics-informed DeepONets. This results in learned basis functions, i.e., the output of the body network in MH-PINN or trunk network in PI-DON, which we store and hold constant for the second optimization step. By doing this, we have evaluated $\mathit{inf_{W^1}}$ in Equation \ref{eq:Knw} because multitask model optimization if over the full \textit{inf} \textit{sup} \textit{inf} and thus the outer \textit{inf} can be taken as an approximate due to the limitations of nonconvex neural network training. The second step is to perform bi-optimization, which competitively trains the basis coefficients between the reference solution and the PIML model solution. Note that our multitask PDE problems are bounded, which is implemented by passing the coefficient variables through a sigmoid function and scaling it to the appropriate parametric interval for the problem. This step can be written as
\begin{align}
\label{eq:metric}
\max_c \min_{W^2} \left| \left| \sum^N_{i=1} c_i \phi_i(\mathbf{x},t) - \sum^M_{i=1} W^2_i \phi_{i}^{W^{1^*}}(\mathbf{x},t)\right| \right|_2 
\end{align}
where we take the norm in Equation \ref{eq:Knw} as the two-norm (i.e., $\left| \left| \cdot \right| \right|_2 = \sqrt{\sum_{i=1}^{D} \left| \cdot \right|^2}$) over the spatiotemporal discretization $(\mathbf{x},t) \in \Omega \times T$, and assume through sufficient optimization the numerically derived \textit{max min} approximates the \textit{sup inf}. This process is provided in Algorithm \ref{alg:metric}, where any optimizer can be used. The size of the network which learns $\phi_i^{W^{1^
*}}$ is arbitrary; we study this choice further in the numerical experiments.

\begin{rmk}While there is no strict rule on which optimization scheme to use, it is important to note that this choice can play a large role in the final result. Different simultaneous optimization schemes result in different training trajectories. We do not expect monotonic convergence of the metric with respect to $n$, i.e., the number of basis functions. In Section \ref{sec:simul_optimization}, we briefly discuss this in the context of PIML. Additionally, the choice of error estimator is also up to the user; here we construct manufactured solutions, but FEM approximations could be used. We discuss this as future work in Sections \ref{sec:simul_optimization} and \ref{sec:conclusions}.
\end{rmk}
\begin{algorithm}[H]
\caption{Algorithm for computing Kolmogorov n-width metric for multitask PIML architectures}
\label{alg:metric}
\begin{algorithmic}
    \REQUIRE{$\mathcal{P}: \mathcal{P} \in$ Error estimator}
    \REQUIRE{$\mathcal{D}: \mathcal{D} \in$ Optimizer}
    \REQUIRE{$\mathcal{L} =$ Physics-informed multitask loss function}
    \REQUIRE{$\mathcal{K}(\mathcal{P}(\cdot)) =$ PIML Kolmogorov n-width (see Equation \ref{eq:Knw})} 
    \STATE{Initialize model parameters $\boldsymbol{\theta} \in \mathbb{R}$}
    \FOR{$n \gets 1$ \TO $N$}
    \STATE{$\boldsymbol{\theta}_{n+1} \gets \mathcal{D}(\mathcal{L}(\mathbf{x}, t; \boldsymbol{\theta}_{n}))$} \hfill $\vartriangleright$ Model optimization
    \ENDFOR
    \STATE{Store learned basis functions $\phi^{W^{1^*}} \gets \phi(W^1_N) \gets \boldsymbol{\theta}_N$ (see Figure \ref{fig:MHPINN+PIDON_diagram})}  
    \STATE{Initialize $\mathcal{K}$ parameters $\{ c, W^2 \}\in \mathbb{R}$}
    \FOR{$m \gets 1$ \TO $M$}
    \STATE{$c_{m+1} \gets \mathcal{D}(\mathcal{K}(\mathcal{P}(\phi^{W^{1^*}}, c_{m}, W^2_{m})))$} \hfill $\vartriangleright \mathcal{K}$ bi-optimization
    \STATE{$W^{2}_{m+1} \gets \mathcal{D}(\mathcal{K}(\mathcal{P}(\phi^{W^{1^*}}, c_{m}, W^2_{m})))$} \hfill $\vartriangleright \mathcal{K}$ bi-optimization
    \ENDFOR
\end{algorithmic}
\end{algorithm}

\subsection{Multitask PIML Regularization}
To perform multitask PIML learning with Kolmogorov n-width regularization, we propose simultaneous tri-optimization. Similar to computing the metric in Equation \ref{eq:metric}, we write our approximation of the Kolmogorov n-width as a \textit{min max min}. This value is then added as a regularizer into the standard physics-informed optimization for the chosen model, such as Equation \ref{eq:MHPINN} for MH-PINNs and Equation \ref{eq:PIDON} for PI-DON.
\begin{align}
\label{eq:regularization}
\min_{W^1} \max_c \min_{W^2} \left| \left| \sum^N_{i=1} c_i \phi_i(\mathbf{x},t) - \sum^M_{i=1} c_i(W^2_i) \phi_{i}(\mathbf{x},t; W^1)\right| \right|_2 
\end{align}
This process is provided in Algorithm \ref{alg:regularization}.
\begin{algorithm}[H]
\caption{Algorithm for training multitask PIML architecture with Kolmogorov n-width regularization}
\label{alg:regularization}
\begin{algorithmic}
    \REQUIRE{$\mathcal{P}: \mathcal{P} \in$ Error estimator}
    \REQUIRE{$\mathcal{D}: \mathcal{D} \in$ Optimizer}
    \REQUIRE{$\mathcal{L} =$ Physics-informed multitask loss function}
    \REQUIRE{$\mathcal{K}(\mathcal{P}(\cdot)) =$ PIML Kolmogorov n-width (see Equation \ref{eq:Knw})} 
    \STATE{Initialize parameters $\{\boldsymbol{\theta},c, W^2\} \in \mathbb{R}$}
    \FOR{$n \gets 1$ \TO $N$}
    \STATE{$\boldsymbol{\theta}_{n+1} \gets \mathcal{D}(\mathcal{L}(\mathbf{x}, t; \boldsymbol{\theta}_{n}) + \mathcal{K}(\mathcal{P}(\phi (W^1(\boldsymbol{\theta}_n)), c_{n}, W^2_{n})))$} \hfill $\vartriangleright$ Model tri-optimization
    \STATE{$c_{n+1} \gets \mathcal{D}(\mathcal{K}(\mathcal{P}(\phi (W^1(\boldsymbol{\theta}_n)), c_{n}, W^2_{n})))$} \hfill $\vartriangleright \mathcal{K}$ tri-optimization
    \STATE{$W^{2}_{n+1} \gets \mathcal{D}(\mathcal{K}(\mathcal{P}(\phi (W^1(\boldsymbol{\theta}_n)), c_{n}, W^2_{n})))$} \hfill $\vartriangleright \mathcal{K}$ tri-optimization
    \ENDFOR
\end{algorithmic}
\end{algorithm}

\begin{figure}[b]
\centering
\includegraphics[width=0.9\textwidth]{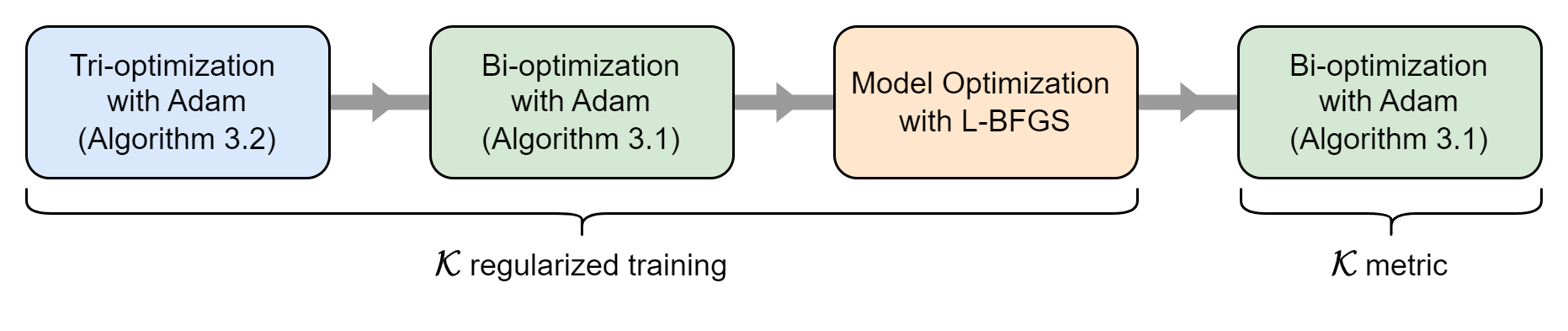}
  \caption{Multistep optimization process used to incorporate L-BFGS training into our Kolmogorov n-width regularization scheme. If L-BFGS training is not necessary, Algorithm \ref{alg:regularization} can be used as is, e.g., with Adam optimizer only. Model optimization (step 3) is architecture dependant and explicitly defined in Equation \ref{eq:MHPINN} for MH-PINNs and Equation \ref{eq:PIDON} for PI-DON. The only difference here is that we are regularizing with the Kolmogorov n-width term, for which the ``most challenging'' problem in the solution manifold has been estimated in the first two steps, and results in the term in Equation \ref{eq:model_optimization_with_regularization}.}
  \label{fig:regFlow}
\end{figure}

In practice, training physics-informed models is commonly done in two optimization stages: Adam \cite{kingma2014adam} and then L-BFGS \cite{liu1989limited}.
Adam is a first-order gradient-based stochastic
optimizer with a low memory requirement. Adam is used in the first stage of training as a warm-up because it has been empirically shown to handle the large jumps in model parameter updates that come with randomly initialized weights and biases of the model. L-BFGS is a quasi-Newton ($1.5$-order) gradient-based optimizer with a low memory requirement. L-BFGS is used in the second stage since it has been empirically shown to converge to much lower physics-informed losses than Adam only. However, it is often unstable in the early stages of training, far away from a local minimum's basin of attraction. Therefore, Adam is used to overcome the initial instability in this method, and L-BFGS is used to refine $\boldsymbol{\theta}$ to achieve orders of magnitude lower loss. Following this logic and based on trial and error, we find that the tri-optimization process in Algorithm \ref{alg:regularization} works with three Adam optimizers. However, when training with L-BFGS as the model optimizer, it becomes unstable and fails due to the updates being made by the other two optimizers. Therefore, to obtain the best model performance, which requires we use L-BFGS, we propose the following optimization steps in Figure \ref{fig:regFlow}. First, we perform Algorithm \ref{alg:regularization} with Adam to approximate the parameter sets $\{\boldsymbol{\theta}, c, W^2\}$. We then hold $\boldsymbol{\theta}$ constant and perform Algorithm \ref{alg:metric} to refine our ``worst possible'' regularization case parameterized by $c$. We then hold $c$ constant ($c^* \gets c$) and perform model optimization with L-BFGS training on $\{\boldsymbol{\theta}, W^2\}$, which is standard to finish physics-informed training as discussed. Model optimization is the processes by which we try to fit our sampled tasks using the model parameterization in Figure \ref{fig:MHPINN+PIDON_diagram}. In terms of the multistep process in Figure \ref{fig:regFlow}, this results in the following regularization term being added: 
\begin{align}
\label{eq:model_optimization_with_regularization}
\left| \left| u(c^*) - \tilde{u}(W^1, W^2) \right| \right|_2
\end{align}
Finally, we perform Algorithm \ref{alg:metric} again with the model basis refined by L-BFGS to compute the final Kolmogorov n-width metric for reporting. Alternatively, if L-BFGS training is unnecessary, Algorithm \ref{alg:regularization} can be used as is for Kolmogorov n-width regularization.

\begin{rmk}
Algorithmic stability, especially in the case of PIML training, can be sensitive to hyperparameter choice. This is particularly true when balancing the loss terms; in our numerical experiments (see Equation \ref{eq:total_loss}), we select hyperparameters that give stable results for the problems herein. However, there are numerous methods toward addressing this issue for PIML that could be added onto our methodology, such as \cite{wang2022and,xiang2022self,mcclenny2020self,subramanian2022adaptive}, and the user should use their own discretion to achieve stable results as it is problem dependant.
\end{rmk}

\subsection{Simultaneous Optimization \& Solution Space Construction}
\label{sec:simul_optimization}
In this section, we briefly discuss different simultaneous optimization schemes that could be used to compute the proposed metric, as well as investigate choices made when constructing the reference solutions during simultaneous optimization. Simultaneous optimization is ubiquitous in other machine learning domains such as Generative Adversarial Networks (GANs) \cite{goodfellow2014generative} and has become prevalent in PIML with the use of self-adaptive weighting schemes \cite{mcclenny2020self,XIANG202211}. Similarly, we must perform simultaneous optimization in Algorithms \ref{alg:metric} \& \ref{alg:regularization}. Given that part of this optimization is finding optimal basis functions from a deep neural network, the overall approximation of the Kolmogorov n-width will be non-convex if done simultaneously. Non-convex min-max optimization, its challenges, and applications have been thoroughly studied here \cite{razaviyayn2020nonconvex}. We choose to use an optimization scheme common for physics-informed training, which are the Adam and L-BFGS optimizers, where the max is taken as the negative loss to perform gradient ascent, similar to self-adaptive PINN weights in \cite{mcclenny2020self}. Another option is to use competitive gradient descent \cite{schafer2019competitive}, which is a robust scheme used to pit two neural network agents against each other in a min-max problem and avoids oscillatory behavior, extended to PINNs here \cite{zeng2022competitive}. However, in our Kolmgorov n-width case, our reference solution does not result from a second neural network. Finally, it is important to keep in mind with all non-convex optimization that the result will be an approximation since the global minimum is not guaranteed and that different realizations of the same process may lead to different results due to different training trajectories, parameter initialization, saddle points, etc. 

Next, we investigate the effect of different choices made during the construction of the solution space. In the context of our numerical results (see Section \ref{sec:numerical_results}), we use a sum of sine functions with variable coefficients to construct 1D Poisson and 2D nonlinear Allen-Cahn solution spaces without a magnitude falloff in higher frequency components. Normally, an elliptic problem would result in a simpler solution space biased toward lower frequencies than the solution we construct here. This limitation on high-energy states is necessary in traditional methods for convergence, which is not the case for neural networks. We make this choice to highlight the effect of neural network activation function choice on the generalizability of the learned basis functions. In this sense, we expect that sine activation functions should not only perform well on the sampled tasks but also generalize better to ``worst-case'' scenarios compared to tanh. Tanh activation functions, in the spirit of the Universal Approximation Theorem \cite{HORNIK1989359} for neural networks, can learn a set of basis functions that fit periodic high-frequency features well. However, this does not guarantee the learned basis will generalize well to unseen high-frequency solutions in the solution space. We observe this numerically and elaborate further in Section \ref{sec:numerical_results}. 

\begin{figure}
\centering
\includegraphics[width=0.9\textwidth]{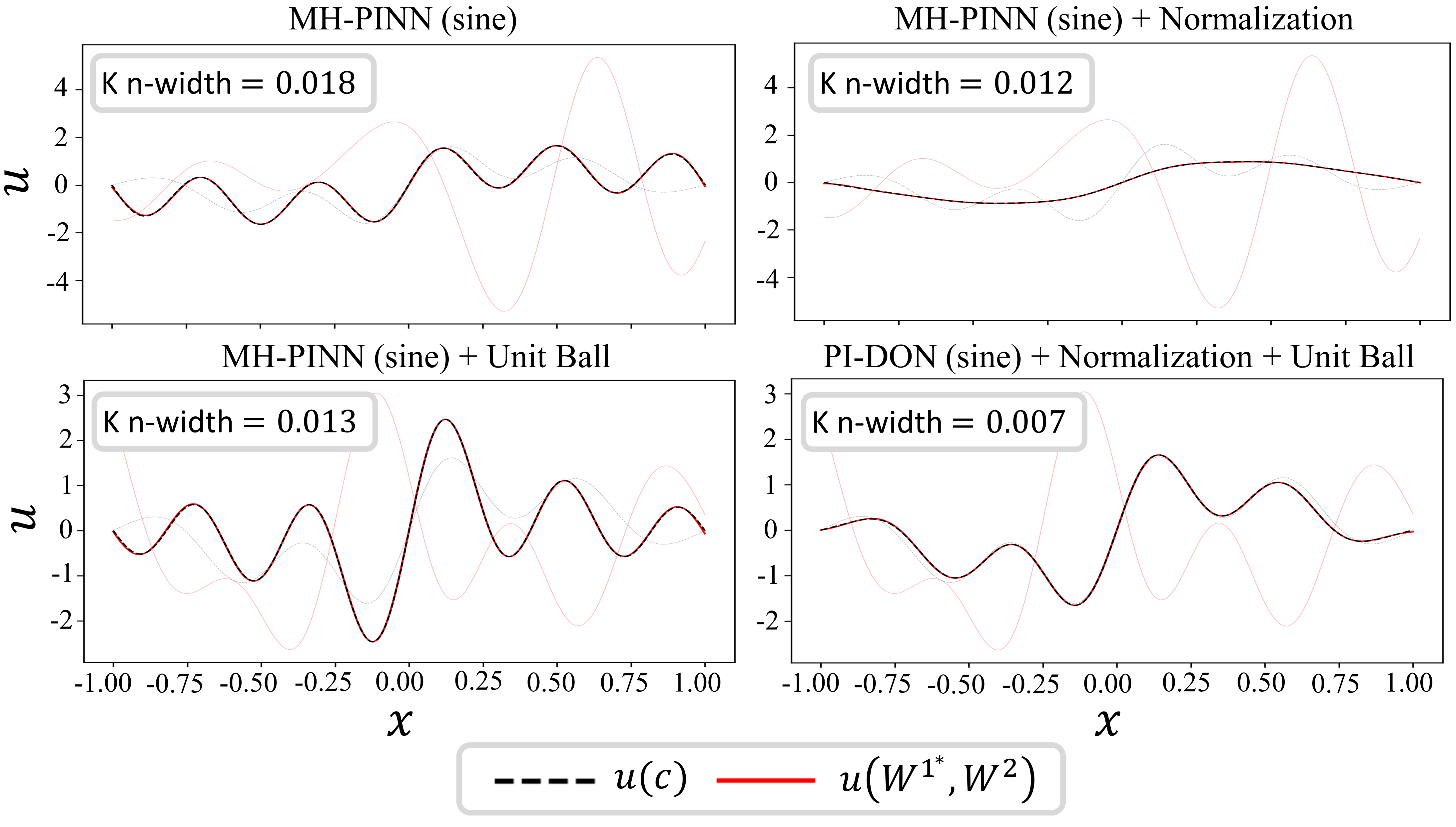}
  \caption{Plots of the Kolmogorov n-width competitive optimization results where the transparent lines represent the start of training and the solid lines represent the end of training.}
  \label{fig:solutionChoice}
\end{figure}

Now, consider we want to rectify our construction of the solution space by adding back the features we would obtain by manufacturing the forcing term and deriving the solution. In the context of the Kolmogorov n-width metric we defined or PIML multitask architectures in Equation \ref{eq:Knw}, we can add normalization with the forcing norm:
\begin{align}
\label{eq:Knw_normalize}
\inf_{W^1} \sup_{c} \inf_{W^2} \frac{\left| \left|u(c) - \tilde{u}(W^1, W^2)\right| \right|_\mathcal{M}} {\left| \left|f(c)\right| \right|_{\mathcal{N}}}, \; c_i \in [a,b].
\end{align}
By including this normalization, we add back the bias toward low frequencies for Laplacian-looking operators. Finally, consider we also want to analyze our choice of bounds for basis function coefficients ($c_i$). We performed our numerical experiments using a unit hypercube, but a unit ball is another common choice. Given these potential choices, we compute the Kolmogorov n-width, using Algorithm \ref{alg:metric} on MH-PINN with sine activation functions and the other settings defined in Section \ref{sec:numerical_results}, shown in Figure \ref{fig:solutionChoice}. Using the normalization in Equation \ref{eq:Knw_normalize} to add back the low energy bias creates a much similar competitive optimization problem, which does not highlight the discrepancy in tanh high-frequency generalization since solutions in that space are eliminated. Additionally, the coefficient unit ball also simplifies the solution space, removing the possibility of outliers such as the edges of the hypercube. Interestingly, the optimization becomes over-constrained by adding the normalization and unit ball constraints, and the coefficients do not change from their initialization, causing the Kolmogorov n-width to be the lowest due to the model basis functions fitting a fixed function, effectively removing the $sup_c$. For these reasons, we do not use those options in our numerical results so we can compare PIML architectures on sufficiently difficult multitask problems. 

In future work, we hope to forego manufactured solutions and use FEM, or other numerical approximation methods, to construct our reference solutions and use the FEM basis functions during simultaneous optimization to compute the PIML Kolmogorov n-widths. However, this poses a few difficulties, such as how to enforce the correct bounds on basis function coefficients for a FEM surrogate concerning the higher-level multitask bounds for the PDE problem. This would require determining the variation each FEM nodal coefficient can have in a numerically efficient way for PDE problems with initial conditions, boundary conditions, forcing functions, etc., that have their own varying bounds that constrain the multitask problem. We hope this step allows for real application of our Kolmogorov n-width method, such as in hypersonic vehicle \cite{shukla2024deep},  offshore structures \cite{cao2024deep} or earthquake localization \cite{haghighat2023novel} applications solved by DeepONets, which would require a solution surrogate. It is important to note that the solution surrogate, such as FEM, does not make training the PIML redundant since PIML has unique features like learning the operator, which can provide real-time inference on unseen tasks and outperform FEM in complex scenarios such as shape optimization \cite{shukla2024deep}. Therefore, the importance of these models by leveraging traditional methods is a critical union to overcome real-world engineering challenges. 

%% file: results.tex
\section{Numerical Experiments}
\label{sec:numerical_results}
In this section, we demonstrate the efficacy of our proposed methods on various multitask PDE problems. The code and data constituting these results are provided.\footnote{The code and data accompanying this manuscript will be made publicly available at \url{https://github.com/mpenwarden/Knw-PIML} after publication} Ground truth solutions are generated using the method of manufactured solutions to derive the correct forcing term. For optimization of all models when using Kolmogorov n-width as a metric (see Algorithm \ref{alg:metric}), we use 1,000 Adam epochs followed by 5,000 L-BFGS epochs for the first model optimization step and then two separately defined Adam optimizers for 5,000 epochs for the bi-optimization step. For optimization of all models when using Kolmogorov n-width as a regularizer (see Algorithm \ref{alg:regularization} \& Figure \ref{fig:regFlow}), we use 1,000 epochs of Adam for the tri-optimization warm-up procedure followed by the same epoch counts in the model optimization and bi-optimization steps as before. The total loss for any given model can be written as
\begin{align}
\label{eq:total_loss}
    \mathcal{L} = \lambda_{\mathcal{R}} \mathcal{L}_{\mathcal{R}} + \lambda_{\mathcal{B}} \mathcal{L}_{\mathcal{B}} + \lambda_{\mathcal{K}} \mathcal{L}_{\mathcal{K}}
\end{align}
where the Lagrangian multipliers are $\lambda_{\mathcal{R}} = 1$ and $\lambda_{\mathcal{B}} = \lambda_{\mathcal{K}} = 10$ which help balance the loss function for optimal training. The terms $\mathcal{L}_{\mathcal{R}}$ and $\mathcal{L}_{\mathcal{B}}$ represent the loss of the PDE residual and boundary condition, respectively (explicitly stated in Equations \ref{eq:MHPINN} \& \ref{eq:PIDON}). The term $\mathcal{L}_{\mathcal{K}}$ is $0$ if unused, otherwise representing the Kolmogorov n-width regularization (see Equation \ref{eq:regularization}). Each neural network (MH-PNN ``body'' or PI-DON ``branch'' \& ``trunk'') is width 20 and depth two unless otherwise stated. Given that this study aims to compare fundamental aspects of different PIML architectures, small networks are sufficient in most cases. However, we also study the effect of varying sizes and training epochs on the computed metrics. Model performance is reported in relative $\ell_2$ error given by
\begin{align*}
    \frac{||u-u_{\bm{\theta}}||_2}{||u||_2}.
\end{align*}
When reporting the Kolmogorov n-width using the 2-norm, we also report it as the relative $\ell_2$ error since this metric has become standardized across PINN literature to report performance and allows for direct comparison with the mean of the multitask problem.

\subsection{1D Poisson equation}
\begin{figure}
\centering
\includegraphics[width=0.9\textwidth]{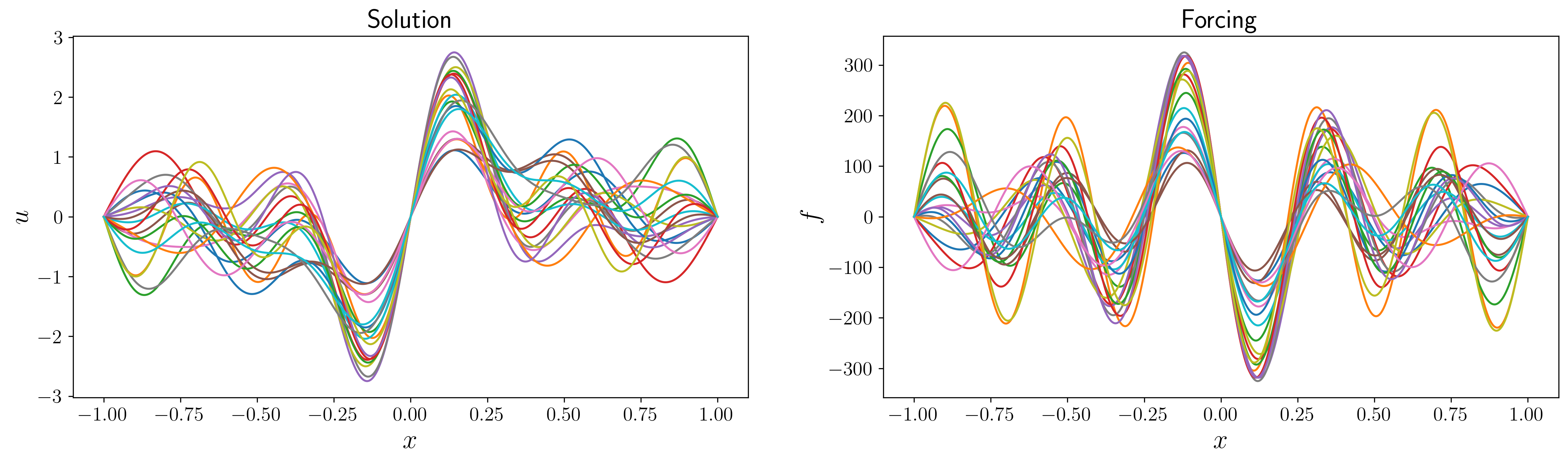}
  \caption{Plot of $20$ 1D Poisson solutions drawn from Equation \ref{eq:poisson_sol} and the derived forcing function respectively.}
  \label{fig:poisson_data}
\end{figure}

Let us consider the following PDE problem
\begin{linenomath}\begin{align*}
& \frac{\partial^2 u}{\partial x^2} = f(x), x\in [-1,1], \\
& u(-1) = u(1) = 0.
\end{align*}\end{linenomath}
To form the dataset, we assume the solution takes the form 
\begin{linenomath}\begin{align}
\label{eq:poisson_sol}
u(x) = \sum^5_{k=1} c_k sin(k x \pi) 
\end{align}\end{linenomath}
with the corresponding forcing term
\begin{linenomath}\begin{align*}
f(x) = -\pi^2 \sum^5_{k=1} k^2 c_k sin(k x \pi) 
\end{align*}\end{linenomath}
where $c_k$ are independent and identically distributed (i.i.d) random variables drawn from the uniform distribution $U(0,1)$. We use a uniform distribution of $512$ residual points for physics-informed training and competitive optimization for the PIML Kolmogorov n-width estimate. The generated dataset is shown in Figure \ref{fig:poisson_data}.

Table \ref{tb:poisson} provides the results of different PIML architectures on the 1D Poisson problem. The (sine) and (tanh) nomenclature denote the type of activation function used in the network representing the basis functions (i.e., the ``body'' and ``trunk'' nets). Adding the $\mathcal{K}$ term denotes that our proposed Kolmogorov n-width regularization was used. The Kolmogorov n-width metric and mean and standard deviation of the sampled multitask problems are reported as the relative $\ell_2$ error for comparison. Note that the results for Kolmogorov n-width converge because the solution coefficients $c_k$ are individually bounded; this is done in the context of machine learning by treating the coefficients as any other neural network learnable parameter, which is then passed through a sigmoid activation function to bound them on the interval $(0,1)$ during optimization. 

Comparing MH-PINN (sine) to MH-PINN (tanh), if one were to report the error over the discretely sampled multitask set, then one would observe the tanh architecture has a reported mean error $3.1\times$ that of the sine architecture. However, when comparing the architectures with the Kolmogorov n-width, the difference is now $16.2\times$ greater. Similarly, the MH-PINN (sine) architecture has a sampled error $4.4\times$ greater than PI-DON (sine), which changes to $12.2\times$ when reported as the Kolmogorov n-width. Furthermore, the Kolmogorov n-width evaluation, representing the lower accuracy bound, is, in both cases, an order of magnitude larger than the discretely sampled distribution of problems. In the regularization case, the sampled error increases, indicating less overfitting of the sampled tasks, which causes poor generalization to ``worst-case'' scenarios in the solution space. This indicates that purely reporting the sampled mean can lead to deceptive conclusions since any given architecture could perform better than another on a subset of problems, but when considering the solution space as a whole, that comparison could change. We claim this is a more appropriate and judicious metric to report as it is not subject to selective task sampling biases.

We also observe little to no increase in computational cost for regularized training because we freeze the main body of the network during competitive optimization in Algorithm \ref{alg:metric} which is used in the multistep regularization process in Figure \ref{fig:regFlow}. The other steps in Figure \ref{fig:regFlow} are also minimal because tri-optimization (Algorithm \ref{alg:regularization}) is only run for 1,000 iterations as an approximation, and L-BFGS, the greatest cost contribution, is necessary with or without regularization. Adding regularization during L-BFGS only adds $20$ new learnable parameters into training.

\begin{rmk}Note that we are comparing MH-PINNs to PI-DONs on the grounds of their learned global basis functions; they are functionally different in that the former is a neural solver and the latter a neural operator \cite{hao2022physics}. We do not comment on their overall use, and we hope this metric can be used to make further improvements and insights on all types of architectures. The same can be said about the choice of activation function; it is not comprehensive but instead illustrative of the need to report robust metrics. \end{rmk}

\begin{figure}
\centering
\includegraphics[width=0.8\textwidth]{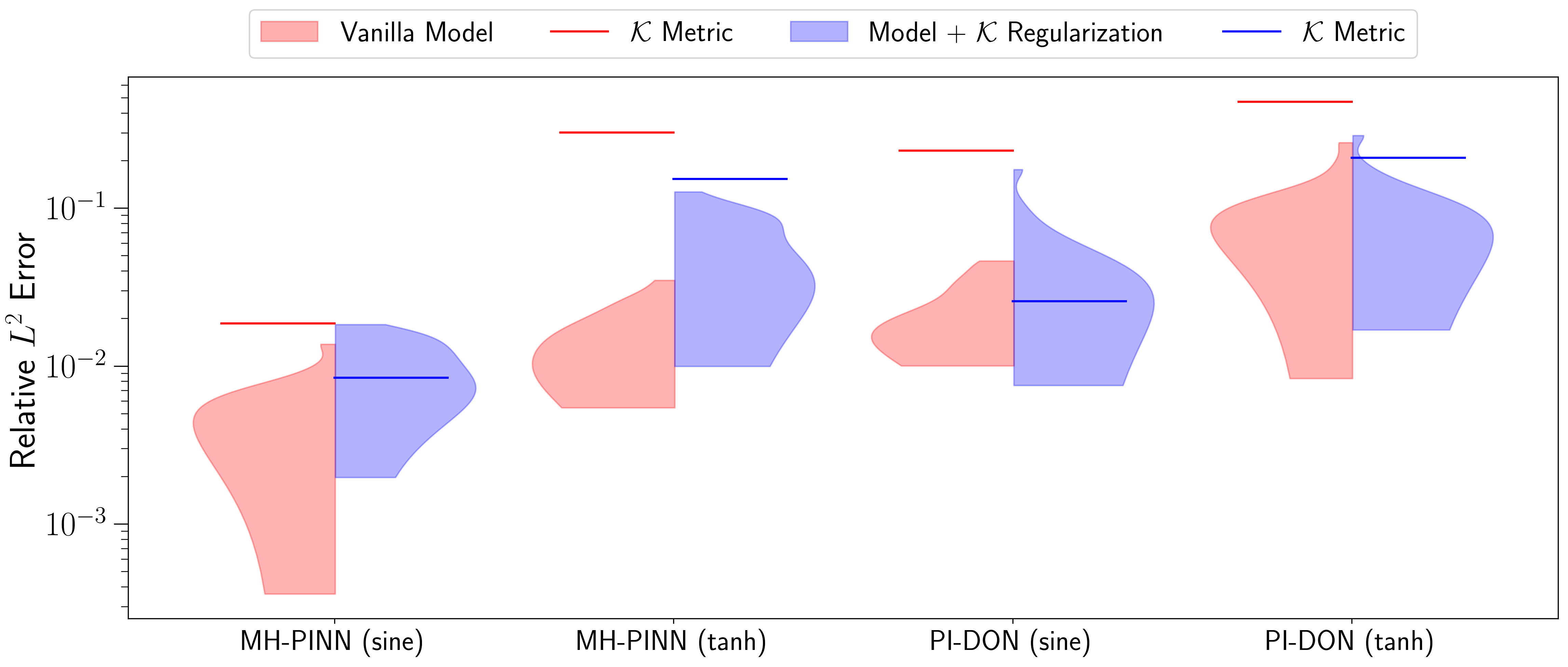}
  \caption{Violin plots showing the distribution of errors for the set of sampled multitask 1D Poisson problems. The resultant Kolmogorov n-width $(\mathcal{K})$ metric is also shown for the same set of basis functions used to solve the sampled problems.}
  \label{fig:poisson_violin}
\end{figure}

\begin{table}
	\centering
    \caption{Table of Kolmogorov n-width values for different PIML architectures for multitask 1D Poisson problem. The Kolmogorov n-width and mean and standard deviation of the sampled multitask problems are reported as the relative $\ell_2$ error for comparison. The runtime is reported as wall-clock-time in seconds over all model training steps.}\label{tb:poisson}
	\begin{tabular}[c]{l | c | c | c }
	\toprule
        Model architecture & Relative Kolmogorov n-width & Sampled Relative $\ell_2$ Error & Runtime (s)\\
        \hline
        MH-PINN (sine) & $0.019 $ & $0.005 \pm  0.003$ & $310$\\
        MH-PINN (tanh) & $0.301 $ & $0.015 \pm  0.008$ & $326$\\
        MH-PINN (sine) + $\mathcal{K}$ & $0.008$ & $0.010 \pm  0.005$ & $304$\\
        MH-PINN (tanh) + $\mathcal{K}$ & $0.153$ & $0.055 \pm  0.003$ & $293$\\
        PI-DON (sine) & $0.232 $ & $0.022 \pm  0.011$ & $346$\\
        PI-DON (tanh) & $0.472 $ & $0.087 \pm  0.057$ & $337$\\
        PI-DON (sine) + $\mathcal{K}$ & $0.026$ & $0.039 \pm  0.038$ & $360$\\
        PI-DON (tanh) + $\mathcal{K}$ & $0.208$ & $0.081 \pm  0.062$ & $357$\\
	\bottomrule
	\end{tabular}
\end{table}
In Figure \ref{fig:poisson_violin}, the results of Table \ref{tb:poisson} are shown as violin plots with the Kolmogorov n-widths denoted for all cases. The violin plots visualize the full distribution of the reported sampled errors in Table \ref{tb:poisson}. We observe that in all cases, using the proposed $\mathcal{K}$ regularization decreased the final Kolmogorov n-width of the trained model. This is particularly true in the cases of the sine activation where the Kolmogorov n-width value is brought in line with the mean of the sample distribution, meaning there is no gap in reporting both metrics. 

While the Kolmogorov n-width represents the lower accuracy bound, it is a numerical approximation conditioned on the assumption of an error estimator to learn the best possible basis coefficients. Note that this differs from the PIML architectures themselves, which are dataless (aside from the boundary conditions) and use physics-informed training to find a solution. In this sense, one can expect that the upper bound of the reported distribution is higher than our numerically computed Kolmogorov n-width due to the training difficulties inherent in PIML. These training challenges and accuracy limitations are well known \cite{wang2023expert, krishnapriyan2021characterizing, PENWARDEN2023112464, WANG2022110768}, and the physics-informed loss does not necessarily correlate to the error. This discrepancy is interesting because if these PIML architectures use their learned basis functions to the best of their ability, they should not result in such high errors, indicating a suboptimal use of their basis. 

\begin{figure}
\centering
\includegraphics[width=0.9\textwidth]{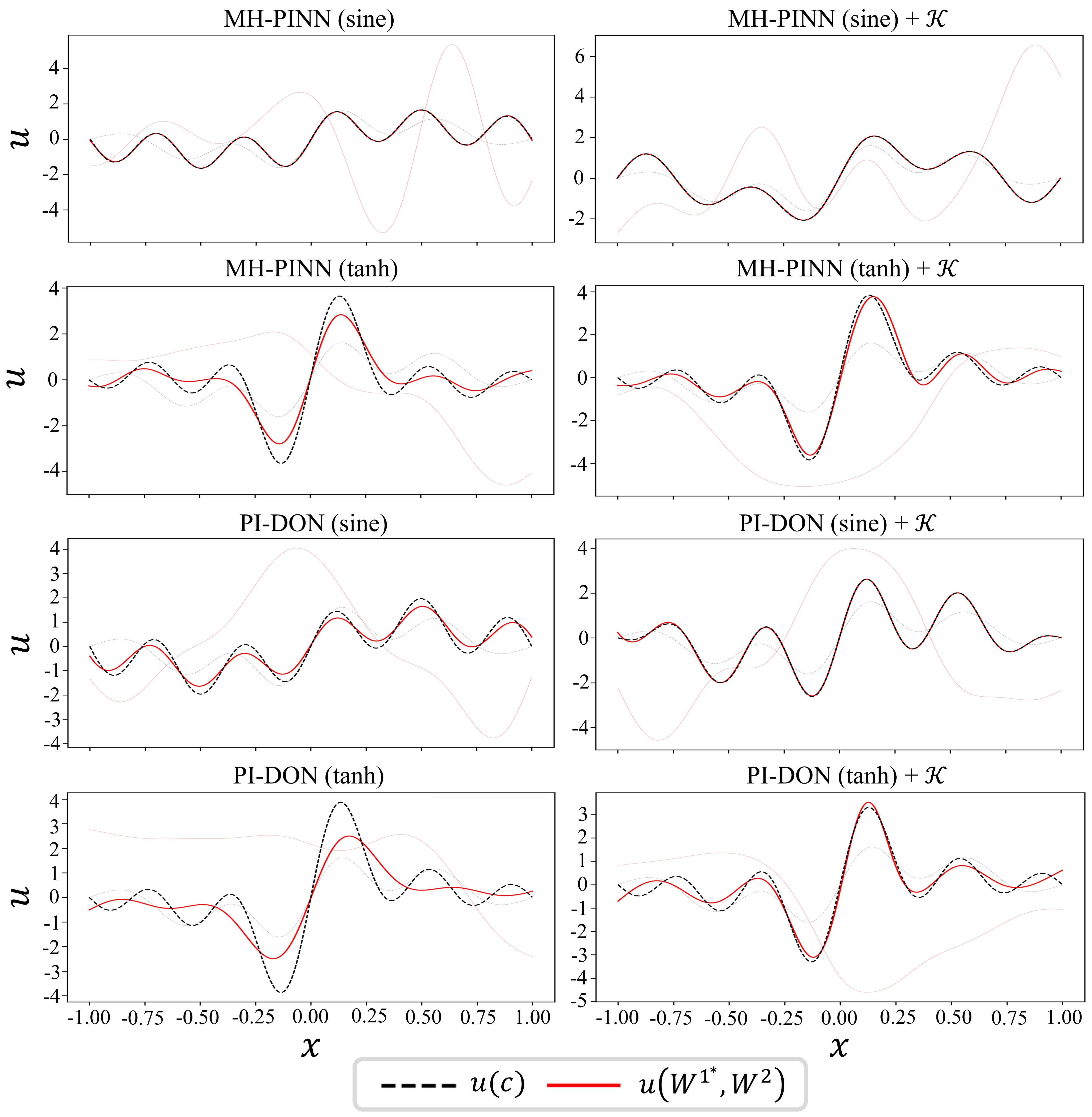}
  \caption{Plots of the Kolmogorov n-width competitive optimization results for the 1D Poisson problem. The \textit{minmax} optimization procedure (shown in Algorithm \ref{alg:metric}) is performed to find the ``worst-case'' scenario in the solution space for a fixed set of learned basis functions $\phi(W^{1^*})$ with the best possible coefficients $(W^2)$.}
  \label{fig:poisson_knw}
\end{figure}

Figure \ref{fig:poisson_knw} shows the Kolmogorov n-width metric results at the end of training, with the more transparent lines representing the beginning of training. Here, we can clearly see the approximation discrepancies made by using sine activation functions over tanh, which struggle to fit the higher frequency features. We can also observe how incorporating the regularization vastly improves the model's approximation of the worst case in the solution space for a given realization of learned basis functions. 

\begin{figure}
\centering
\includegraphics[width=0.65\textwidth]{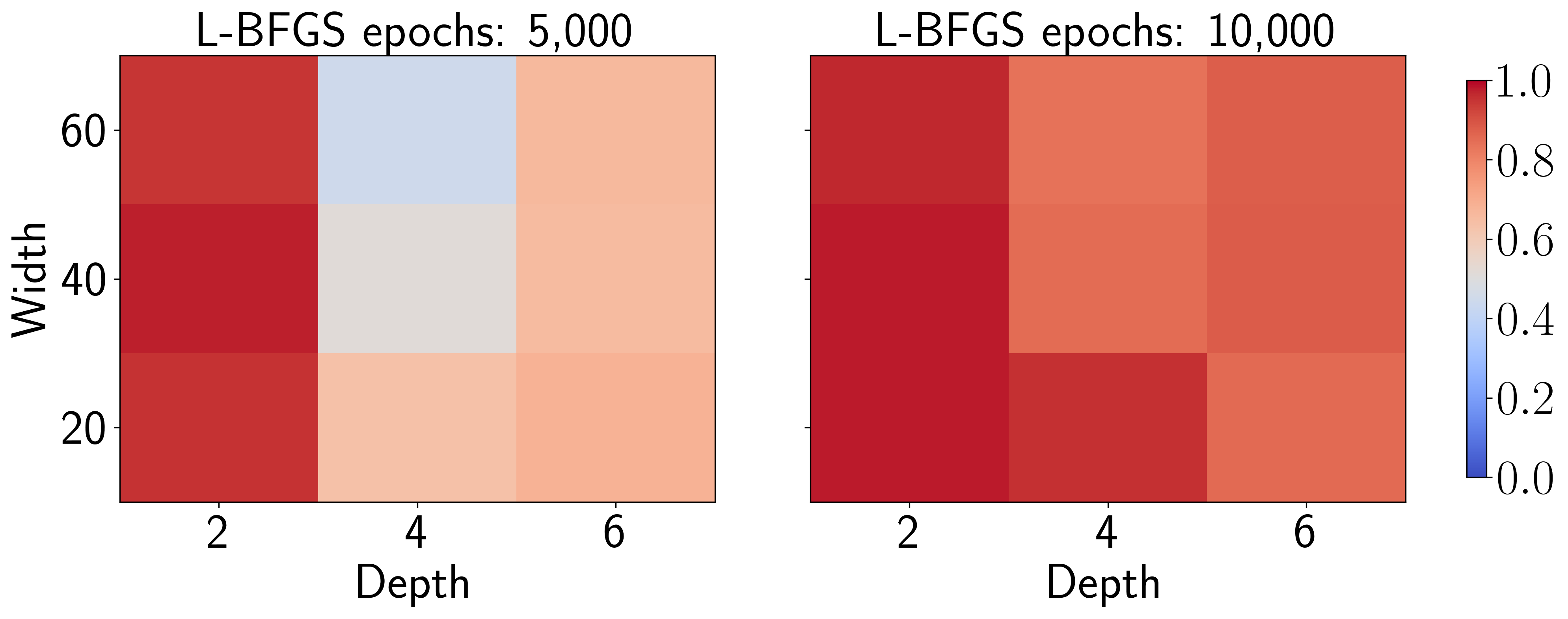}
  \caption{Heat map of the relative difference between the Kolmogorov n-width metric and mean of the sampled multitask problem metric where a value of $0$ represents no reporting discrepancy.}
  \label{fig:poisson_architecture}
\end{figure}

Finally, we study the effect of architecture and optimization criteria choices on the observed gap between reporting the sampled mean versus the Kolmogorov n-width metrics. All models are MH-PINN (tanh) without regularization. In Figure \ref{fig:poisson_architecture}, the relative difference, i.e. $\frac{\mathcal{K}_{\ell_2}-\overline{S}_{\ell_2}} {\mathcal{K}_{\ell_2}}$, is reported for different widths, depths, and optimization epochs (for the L-BFGS step). Note that a value of $0$ indicates no discrepancy between the mean of the sampled multitask error ($\overline{S}_{\ell_2}$) and the Kolmogorov n-width ($\mathcal{K}_{\ell_2}$). The relative difference is clearly more dependent on the depth than the width, which appears almost invariant for shallow networks. 

The lowest value found was $0.45$ for depth = $4$, width = $60$, and epochs = $5,000$; however, this is still around a $2\times$ difference of the metrics. Furthermore, we observe that more training epochs cause the discrepancy to increase. This implies that the lower values from the larger networks with fewer epochs are due to undertraining, which causes the sampled error and, thus, also the ratio to be lower. This causes the model to overfit the sampled tasks even more, given its greater expressivity, and perform worse in terms of Kolmogorov n-width. This study validates that regardless of network choices, a significant gap in Kolmogorov n-width and sampled error is still observed and that it is important to consider the entire solution space instead of potentially selectively sampled tasks with any sized network. 

\subsection{2D nonlinear Allen-Cahn equation}
\label{ssec:allen_cahn}
\begin{figure}[H]
\centering 
\includegraphics[width=0.9\textwidth]{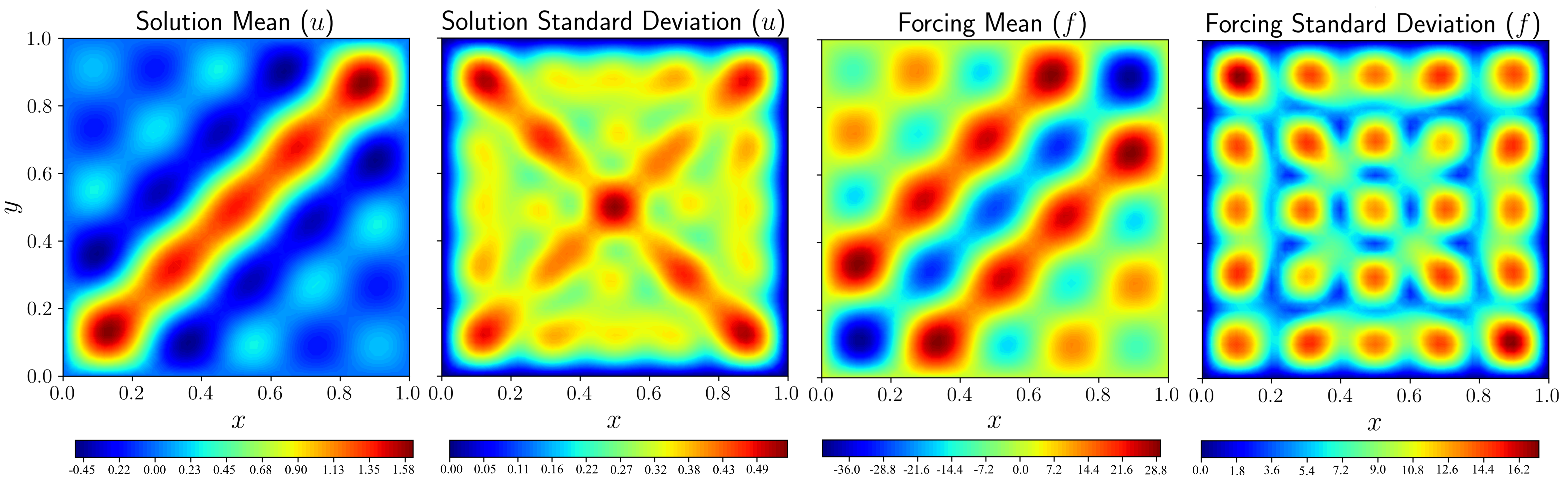}
  \caption{Plot of $20$ 2D nonlinear Allen-Cahn solutions drawn from Equation \ref{eq:allencahn_sol} and the derived forcing functions respectively.}
  \label{fig:allencahn_data}
\end{figure}

Let us consider the following stationary PDE problem
\begin{linenomath}\begin{align*}
& \lambda \left( \frac{\partial^2 u}{\partial x^2} + \frac{\partial^2 u}{\partial y^2} \right) + u(u^2-1) = f(x,y), (x,y) \in [0,1] \times [0,1], \\
& u(x,0) = u(x,1) = u(0,y) = u(1,y) = 0,
\end{align*}\end{linenomath}
where $\lambda=0.1$ and the solution $u(x,y)$ takes the form 
\begin{linenomath}\begin{align}
\label{eq:allencahn_sol}
u(x,y) = \sum^5_{k=1} c_k sin(k x \pi) sin(k y \pi)
\end{align}\end{linenomath}
where $c_k$ are (i.i.d) random variables drawn from the uniform distribution $U(0,1)$. We use a uniform grid of size $51 \times 51$ to evaluate the residual points for physics-informed training and competitive optimization for the PIML Kolmogorov n-width estimate. The generated dataset is shown in Figure \ref{fig:allencahn_data}.

\begin{rmk}
    For ease of implementation of competitively training the solution and learned basis functions to compute the Kolmogorov n-width and to highlight different architecture choices such as sine versus tanh activation functions, we have chosen to construct $u(\cdot)$ out of sine functions instead of $f(\cdot)$. While this is a trivial change in the linear 1D Poisson problem, we acknowledge that this, in some sense, simplifies the nonlinear problem since we would expect a more complex set of solution basis functions if $f(\cdot)$ were constructed out of a sum of sine functions. Our main goal is to showcase our proposed metric to provide insight and benchmarking for PIML architectures, not solve the most challenging problem. In future work, we hope to study more complex constructions of the solution basis, such as using FEM as the reference solution and basis instead of the method of manufactured solutions.  
\end{rmk}

Table \ref{tb:allen-cahn} provides the final results of all considered architectures for the 2D nonlinear Allen-Cahn Problem. Overall, we observe similar trends in the gap between the Kolmogorov n-width metric and the sampled multitask mean. For example, while the sampled means are almost equivalent for MH-PINN (sine) and (tanh), the Kolmogorov n-width is almost five times larger. This indicates our method should be invariant to the spatial dimension, at least to the extent that the Kolmogorov n-width still indicates the models overfit the set of discrete problems regardless of dimension. Interestingly, MH-PINN (sine) has no gap between the two metrics, indicating a good set of learned basis functions. We also observe that unlike MH-PINN (sine), PI-DON (sine) suffers very bad generalizability despite having the advantage of sine activation functions and an incredibly low sampled mean error. This feature is also present in the 1D Poisson results, where the PI-DON architecture fails to generalize well to the worst case despite achieving similar performance on the sampled tasks. It could be the case that the additional complexity of the ``branch'' network, which allows the PI-DON to be a neural operator and predict new solutions without additional training, causes poor generalization of the learned basis functions, which is masked when only reporting the sampled task error instead of the Kolmogorov n-width. These kinds of insights can not only be used to compare competing architectures but also show where improvements can be made, for example, to the generalizability of the ``trunk'' net in PI-DONs. Finally, as in the 1D case, we still note that our Kolmogorov n-width regularization procedure improves the metric in all cases. Explicit visualizations of the learned basis functions can be found in \ref{appendix:basis}.

\begin{table}
	\centering
    \caption{Table of Kolmogorov n-width values for different PIML architectures for multitask 2D nonlinear Allen-Cahn problem. The Kolmogorov n-width and mean and standard deviation of the sampled multitask problems are reported as the relative $\ell_2$ error for comparison. The runtime is reported as wall-clock-time in seconds over all model training steps.} \label{tb:allen-cahn}
	\begin{tabular}[c]{l | c | c | c }
	\toprule
        Model settings & Relative Kolmogorov n-width & Sampled Relative $\ell_2$ Error & Runtime (s)\\
        \hline
        MH-PINN (sine) & $0.064 $ & $0.063 \pm  0.044$ & $992$\\
        MH-PINN (tanh) & $0.397 $ & $0.085 \pm  0.027$ & $924$\\
        MH-PINN (sine) + $\mathcal{K}$ & $0.060$ & $0.092 \pm  0.043$ & $1007$\\
        MH-PINN (tanh) + $\mathcal{K}$ & $0.098$ & $0.249 \pm  0.120$ & $1015$\\
        PI-DON (sine) & $0.347 $ & $0.020 \pm  0.009$ & $1006$\\
        PI-DON (tanh) & $0.652 $ & $0.127 \pm  0.027$ & $963$\\
        PI-DON (sine) + $\mathcal{K}$ & $0.169$ & $0.065 \pm  0.028$ & $1107$\\
        PI-DON (tanh) + $\mathcal{K}$ & $0.217$ & $0.193 \pm  0.049$ & $1013$\\
	\bottomrule
	\end{tabular}
\end{table} 

\begin{figure}
\centering
\includegraphics[width=0.8\textwidth]{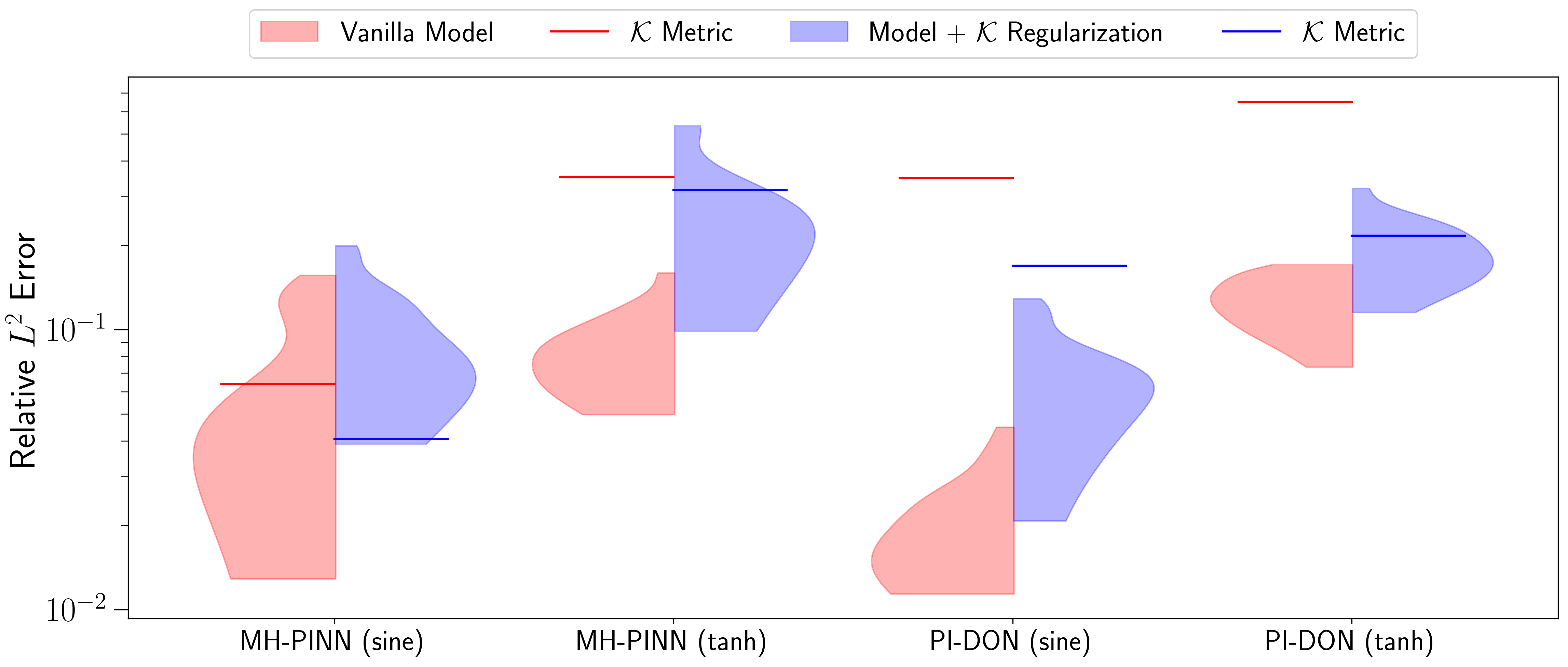}
  \caption{Violin plots showing the distribution of errors for the set of sampled multitask 2D nonlinear Allen-Cahn problems. The resultant Kolmogorov n-width $(\mathcal{K})$ metric is also shown for the same set of basis functions used to solve the sampled problems.}
  \label{fig:allencahn_violin}
\end{figure}

In Figure \ref{fig:allencahn_violin}, the results of Table \ref{tb:allen-cahn} are shown as violin plots for the 2D nonlinear Allen-Cahn problem. Similar to the 1D Poisson problem, we see a large gap between the sampled distribution and the Kolmogorov n-width. However, in the 2D case, the PI-DON architectures (both sine \& tanh) perform much better on the sampled tasks than in 1D, almost equivalent to the MH-PINN distributions. This would imply the models learned a similar set of basis functions given their similar performance. However, we can see that in terms of the Kolmogorov n-width, despite achieving similar performance, the basis are quite different. 

\begin{figure}
\centering
\includegraphics[width=0.9\textwidth]{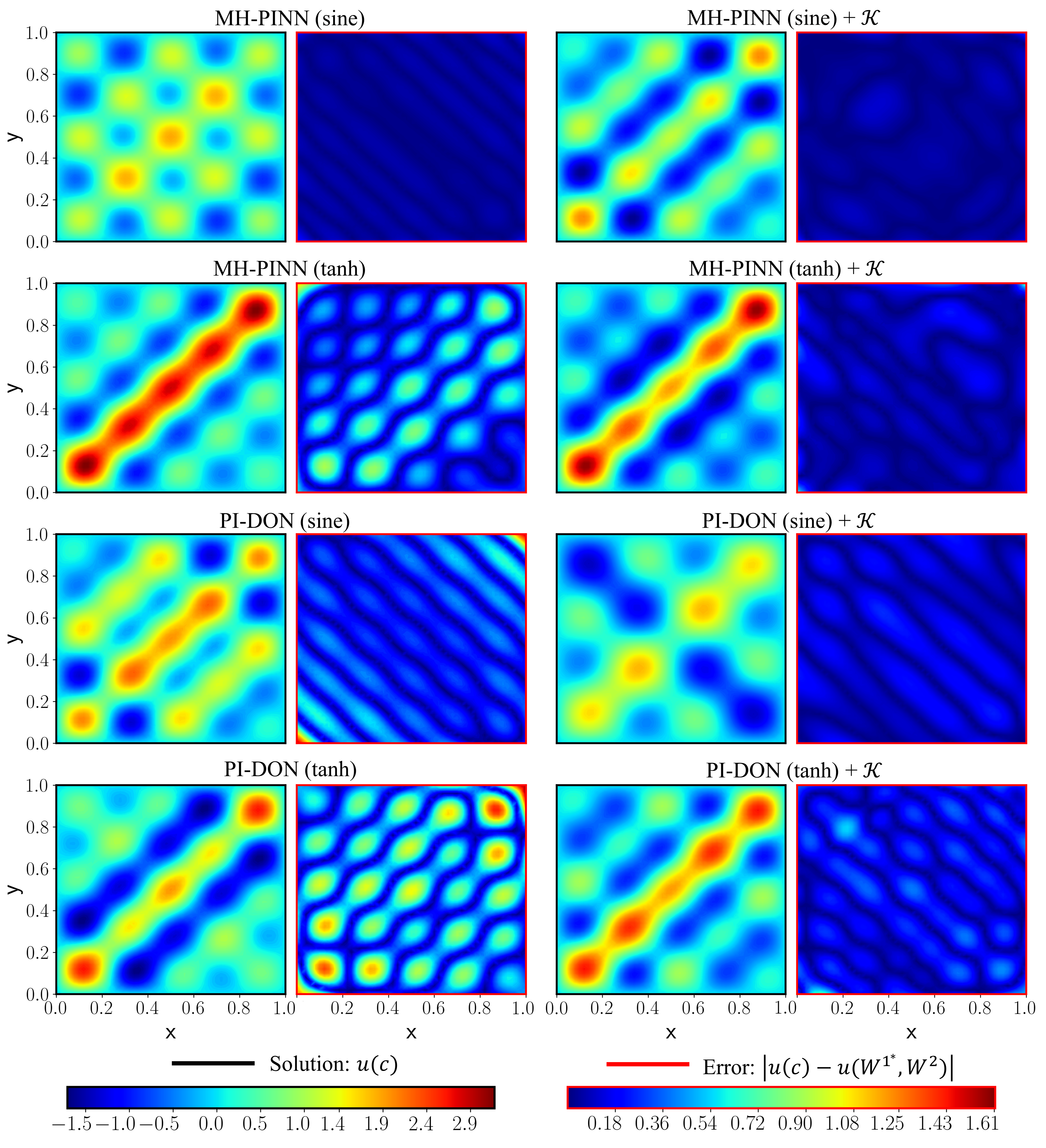}
  \caption{Plots of the Kolmogorov n-width competitive optimization results for the 2D nonlinear Allen-Cahn problem. The \textit{minmax} optimization procedure (shown in Algorithm \ref{alg:metric}) is performed to find the ``worst-case'' scenario in the solution space for a fixed set of learned basis functions $\phi(W^{1^*})$ with the best possible coefficients $(W^2)$.}
  \label{fig:allencahn_knw}
\end{figure}

\begin{figure}
\centering
\includegraphics[width=0.8\textwidth]{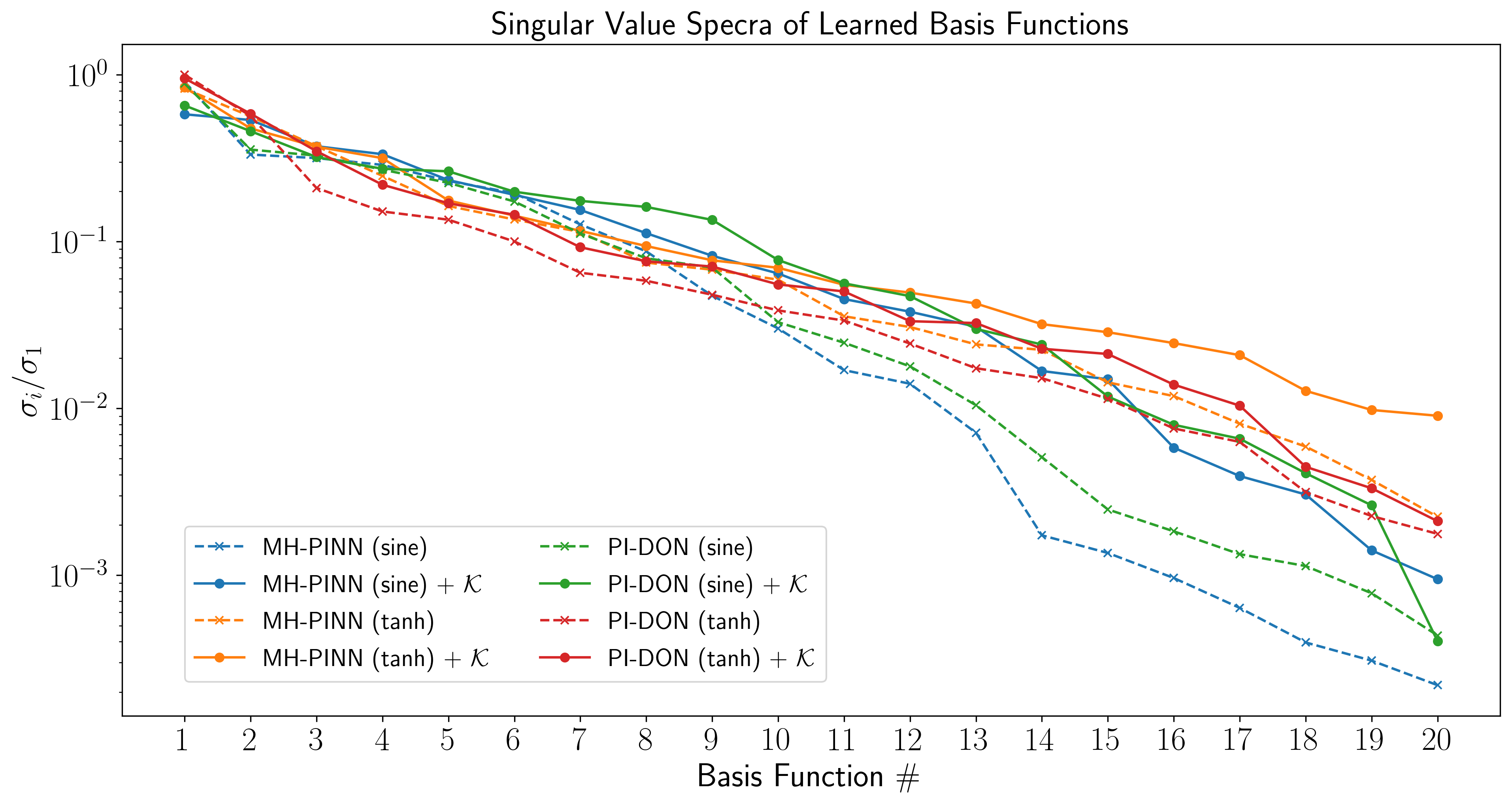}
  \caption{Plots of the rate of decay of (normalized) singular values of the set of learned basis functions. The results are computed by performing singular value decomposition (SVD) on the vectorized set of 2D basis functions where the matrix is of shape: \textbf{neurons in the final layer of the $\boldsymbol{\phi}$ network} $\times$ \textbf{vec(spatial discretization)}, in our case $\mathbf{20} \times \mathbf{2601}$.}
  \label{fig:allencahn_svd}
\end{figure}

Figure \ref{fig:allencahn_knw} shows the Kolmogorov n-width metric results at the end of training. The black outlined plots are the ``worst-case'' scenario, and the red outlined plots are the point-wise error between the ``worst-case'' scenario and the best approximation made with the learned model basis functions. Similar to the 1D case, we also observe that the tanh activation function models struggle to fit higher-frequency features. Interestingly, the point-wise error plots using the regularization method show more complex features than the standard periodic errors without regularization. This, in fact, is indicative that their learned basis functions are richer, which can be seen when plotting, and we further study in Figure \ref{fig:allencahn_svd}.  

Finally, we perform a study of the learned 2D basis functions for the nonlinear Allen-Cahn problem to help quantify and explain the differences observed in the metrics. Instead of manually plotting and searching for visual differences and trends between the output of final layer neurons in the network representing $\phi$, we seek to quantify them with their singular value spectra. In \cite{mojgani2022lagrangian}, the authors propose to use the decay rate of singular values of temporal snapshots of the solution can be used as an \textit{a priori} guideline for the complexity of PINN training.  In a similar vein of investigation, we instead look at the decay rate of singular values of the learned basis functions instead of solution snapshots. This gives us insight into the richness of the learned basis for different architecture and training schemes, such as our regularization procedure. We observe that in all cases, the $\mathcal{K}$ regularization procedure forces the learned basis to become richer and more expressive to fit the features in the ``worst-case'' scenario, which, in combination with the prior discussions on the metric, results in a more generalizable architecture. Interestingly, the sine activation function models have a higher rate of decay than tanh despite performing better or equal in terms of sampled and n-width error. This indicates it is not just about expressivity but also about learning the right set of expressive basis, for which sine, in this case, is expected to perform better in terms of Kolmogorov n-width, given our solution space is constructed from sine functions. This also implies orthogonality does not guarantee generalization in the sense of Kolmogorov n-widths since a completely orthogonal set of basis would have no singular value decay, but tanh, which has less decay, performs worse than sine. In future work, we hope to investigate the effect of enforcing orthogonality of PIML basis functions under the lens of Kolmogorov n-width and singular value spectra to understand the nature of physics-informed training better. We also hope to provide a more comprehensive study of existing PIML training modifications and their effect on the learned basis functions, such as Fourier feature embeddings \cite{dong2021method} or adaptive activation functions \cite{jagtap2020adaptive}.

\begin{rmk}
While the authors of \cite{mojgani2022lagrangian} claim the singular values of snapshots represent the Kolmogorov n-width, it does not say anything about the approximating functions since they only look at the known solution space, not the model approximation space. In our work, we are fully estimating the \textit{inf} \textit{sup} \textit{inf} to obtain our metric. 
\end{rmk}

%% file: summary.tex
\section{Conclusions}
\label{sec:conclusions}
We have introduced a novel optimization procedure for computing the Kolmogorov n-width, a measure of the effectiveness of approximating functions, for analyzing multitask PIML architectures. We have found that multitask PIML models overfit the discrete training tasks which can result in poor generalization to ``worst-case'' scenarios that exist in the entire task family. This can also lead to a form of selective sampling, where reporting the sampled task errors can result in misleading conclusions about the efficacy of the model and its learned basis functions. We objectively compare and benchmark different multitask PIML architectures using the proposed metric. We also find that neural network choices such as activation function have dramatic performance effects and spectral bias only seen under the lens of Kolmogorov n-widths. We also successfully propose and implement a novel regularization scheme using Kolmogorov n-widths to condition the model against overfitting and improve the generalization of the learned basis functions when solving multitask PDE problems. Future investigations will explore using different error estimators, e.g., constructing the reference solution with FEM basis functions to compute the Kolmogorov n-width as a step closer to real-world applications. Without this step, more cases would illustrate a similar outcome, that users should be more cautious when reporting performance metrics. With this extension, we hope that experimenting with more cases such as time-dependent PDEs, Darcy flows, high-dimensional PDEs, etc, may yield more insights. We also hope to comprehensively study the effect existing PIML architecture modifications like adaptive activation functions or Fourier feature embeddings have on the learned basis functions and to benchmark a wider array of problems and architectures.

%% file: appendix.tex
\appendix
\section{Symbols \& Notations}

\begin{table}[H]
	\centering
 \caption{Symbols \& Notations}
	\begin{tabular}[c]{l l}
    \hline
    $u(\cdot)$ & PDE solution \\
    $\phi_i$ & Basis Functions \\
    $c_i$ & Basis Coefficients \\
    $\phi_i(W^1)$ & Network Parameterization of Basis Functions \\
    $c_i(W^2)$ & Network Parameterization of Basis Coefficients \\
    $\boldsymbol{\theta}$ & Learnable network parameters \\
    $\boldsymbol{\theta}^*$ & Optimized network parameters \\
    $u_{\boldsymbol{\theta}}(\cdot)$ & PINN PDE prediction \\
    $\mathcal{R}(\cdot)$ & PDE residual \\
    $\mathcal{F}(\cdot)$ & Nonlinear operator \\
    $\mathcal{S}(\cdot)$ & Source function \\
    $\mathcal{B}(\cdot)$ & Boundary/initial condition \\
    $\mathcal{L}(\cdot)$ & Loss Function \\
    $\mathcal{K}(\cdot)$ & Kolmogorov n-width \\
    $\Omega$ & Spatial domain of interest \\
    $T$ & Temporal domain of interest \\
    $\mathbf{x}$ & Spatial value, $\mathbf{x} \in \Omega$ \\
    $t$  & Temporal value, $t \in T$ \\
    \hline
	\end{tabular}
\end{table}

\section{Learned basis functions for the 2D nonlinear Allen-Cahn problem}
\label{appendix:basis}
\begin{figure}
\centering
\includegraphics[width=1\textwidth]{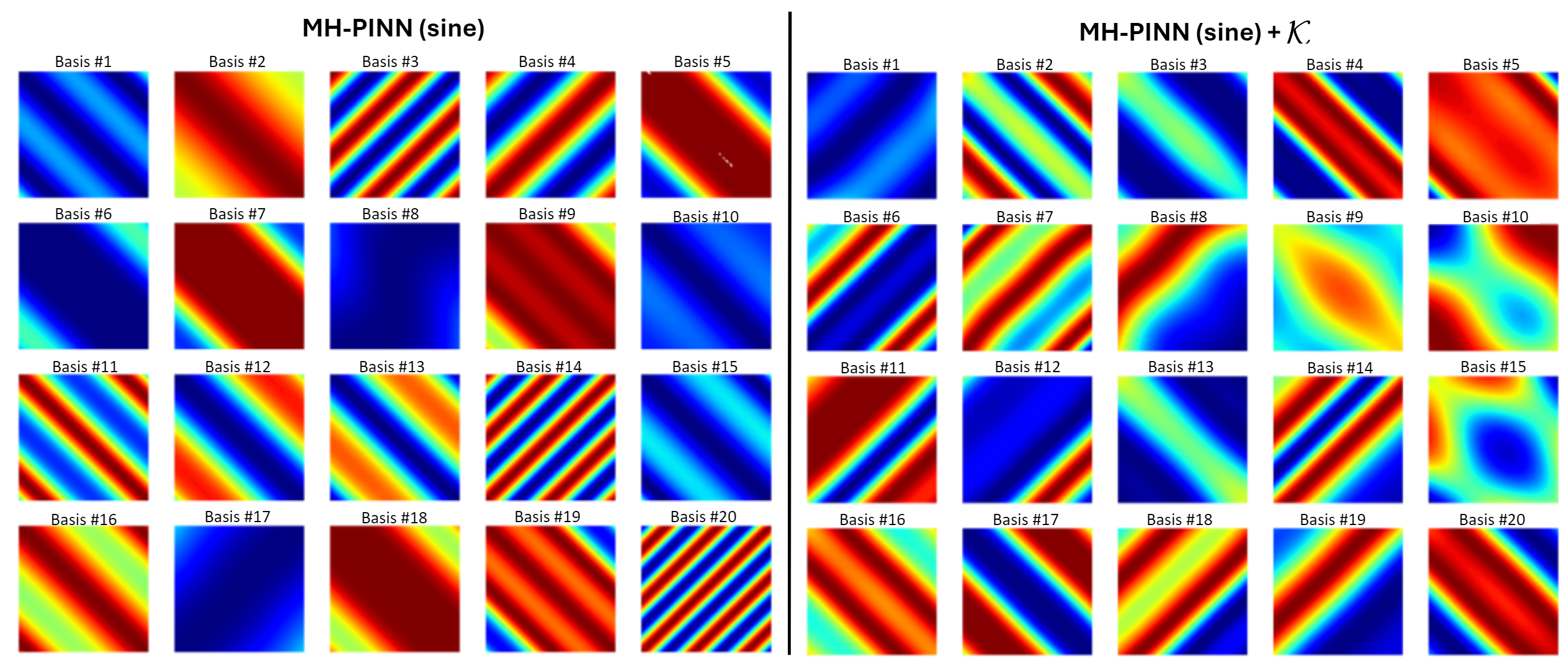}
  \caption{Plot of the learned basis functions at the end of model training for MH-PINN (sine) architecture with and without Kolmogorov n-width regularization. When adding $\mathcal{K}$ regularization, the learned basis is more expressive and more generalizable.}
  \label{fig:MH-PINN_sine_basis}
\end{figure}

\begin{figure}
\centering
\includegraphics[width=1\textwidth]{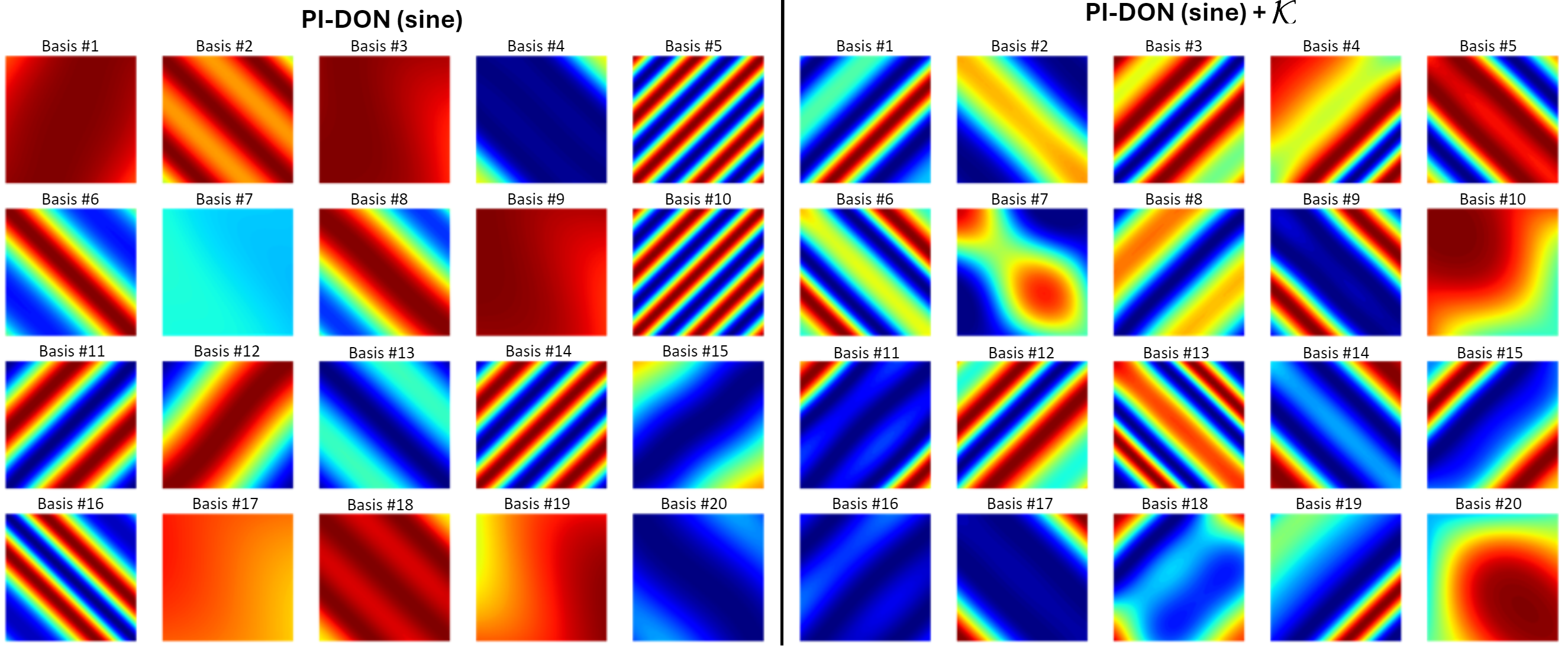}
  \caption{Plot of the learned basis functions at the end of model training for PI-DON (sine) architecture with and without Kolmogorov n-width regularization. When adding $\mathcal{K}$ regularization, the learned basis is more expressive and more generalizable.}
  \label{fig:PI-DON_sine_basis}
\end{figure}

\begin{figure}
\centering
\includegraphics[width=1\textwidth]{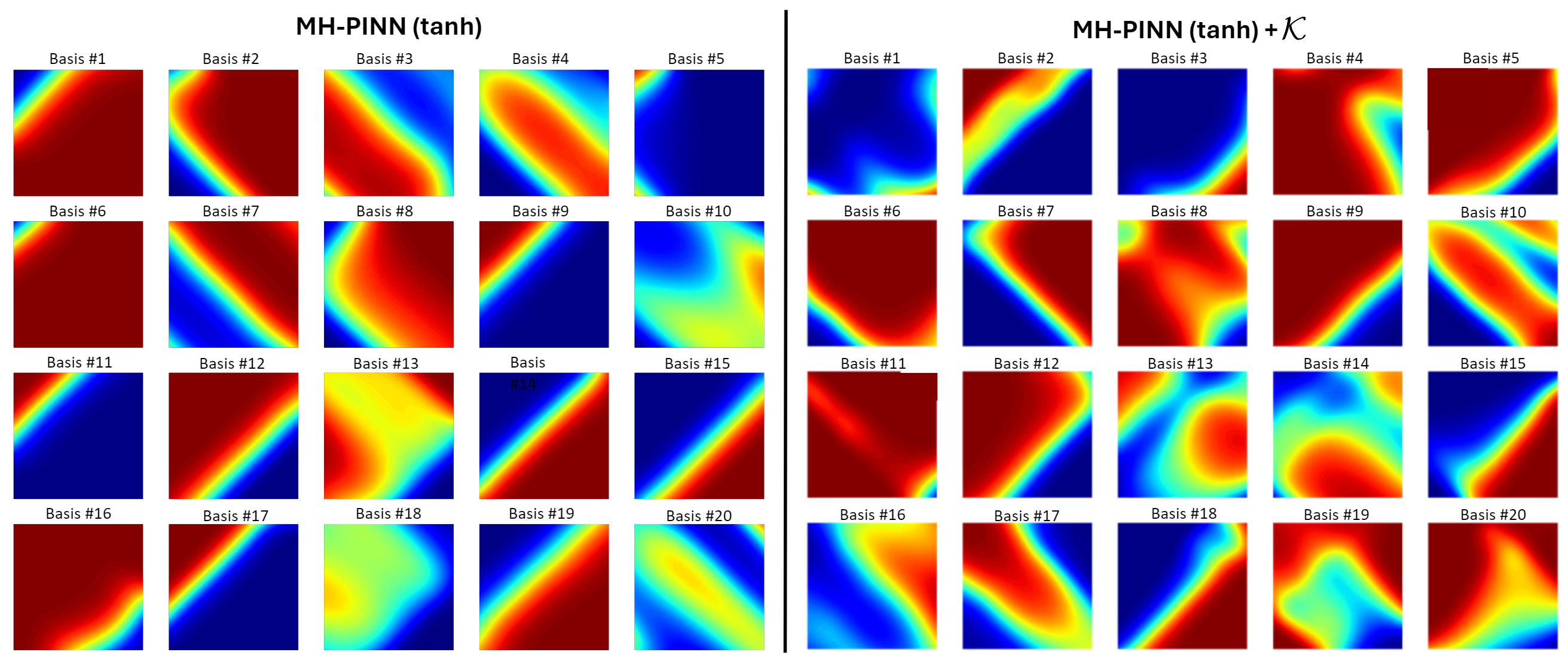}
  \caption{Plot of the learned basis functions at the end of model training for MH-PINN (tanh) architecture with and without Kolmogorov n-width regularization. When adding $\mathcal{K}$ regularization, the learned basis is more expressive and more generalizable.}
  \label{fig:MH-PINN_tanh_basis}
\end{figure}

\begin{figure}
\centering
\includegraphics[width=1\textwidth]{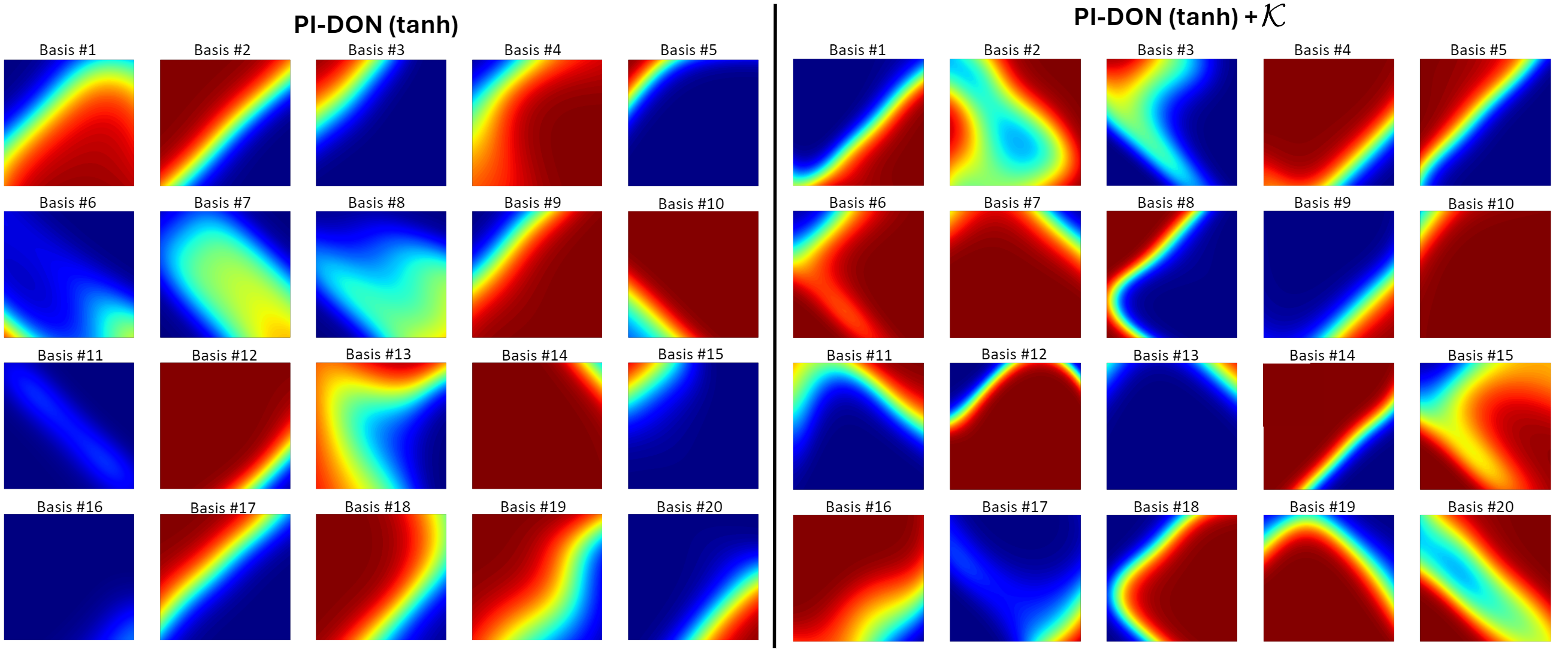}
  \caption{Plot of the learned basis functions at the end of model training for PI-DON (tanh) architecture with and without Kolmogorov n-width regularization. When adding $\mathcal{K}$ regularization, the learned basis is more expressive and more generalizable.}
  \label{fig:PI-DON_tanh_basis}
\end{figure}

In this section, we visualize the learned basis functions of the models trained in Section \ref{ssec:allen_cahn} on 2D nonlinear Allen-Cahn problem. In Figures \ref{fig:MH-PINN_sine_basis} - \ref{fig:PI-DON_tanh_basis}, the output (basis function) of each neuron in the final layer of the body network for the respective architectures are plotted as a function of space $(x,y)$. While the numerical improvements of regularization in terms of the final Kolmogorov n-width metric are evident in Table \ref{tb:allen-cahn}, we find these additional plots add to the intuition and confidence of its effect. For example, looking at PI-DON (sine) in Figure \ref{fig:PI-DON_sine_basis}, the basis functions \#5 and \#10 are practically the same which is a redundant feature. There are also other basis functions such as \#1, 3, 9, and 17 which have very little variation and don't contribute to being able to represent the full space of possible solutions. In contrast, PI-DON (sin) + $\mathcal{K}$ clearly has all unique basis functions. The basis functions for sine activations are also clearly more complex, representing non-diagonal features, which without regularization does not happen. It is clear the learned basis functions are phenomenologically different with and without regularization, which is numerically backed up in the analysis in Figure \ref{fig:allencahn_svd} in the main text and now visualized here.

